\def\year{2022}\relax
\documentclass[letterpaper]{article} 
\usepackage{aaai22}  
\usepackage{times}  
\usepackage{helvet}  
\usepackage{courier}  
\usepackage[hyphens]{url}  
\usepackage{graphicx} 
\usepackage{natbib}  
\usepackage{caption} 
\DeclareCaptionStyle{ruled}{labelfont=normalfont,labelsep=colon,strut=off} 
\usepackage{algorithm}
\usepackage{algorithmic}

\usepackage{mathrsfs}
\usepackage{amssymb}
\usepackage{bm}
\usepackage{amsmath}
\usepackage{dsfont}
\usepackage{booktabs}
\usepackage{epstopdf}
\usepackage{multirow}
\usepackage{comment}
\usepackage[table,xcdraw]{xcolor}
\usepackage{cite}
\usepackage{endnotes}

\usepackage{newfloat}
\usepackage{listings}
\floatstyle{ruled}
\newfloat{listing}{tb}{lst}{}
\floatname{listing}{Listing}
\title{Deep Graph Clustering via Dual Correlation Reduction}
\author{Yue Liu,\textsuperscript{\rm 1}\footnote{First author with equal contribution} Wenxuan Tu,\textsuperscript{\rm 1}\footnotemark[1] Sihang Zhou,\textsuperscript{\rm 2} Xinwang Liu,\textsuperscript{\rm 1}\footnote{Corresponding author} 

Linxuan Song,\textsuperscript{\rm 1} Xihong Yang,\textsuperscript{\rm 1} En Zhu\textsuperscript{\rm 1}}
\affiliations{
    \textsuperscript{\rm 1}College of Computer, National University of Defense Technology, Changsha, China\\
    \textsuperscript{\rm 2}College of Intelligence Science and Technology, National University of Defense Technology, Changsha, China\\
    
    \{yueliu, twx, xinwangliu, yangxihong, enzhu\}@nudt.edu.cn, sihangjoe@gmail.com, slxnatavidad@163.com
}

\usepackage{bibentry}
\usepackage{hyperref}
\hypersetup{
    colorlinks=false,
    linkcolor=blue,
    filecolor=blue,      
    urlcolor=blue,
    citecolor=cyan,
}

\begin{document}

\maketitle

\begin{abstract}

Deep graph clustering, which aims to reveal the underlying graph structure and divide the nodes into different groups, has attracted intensive attention in recent years. However, we observe that, in the process of node encoding, existing methods suffer from representation collapse which tends to map all data into the same representation. Consequently, the discriminative capability of the node representation is limited, leading to unsatisfied clustering performance. To address this issue, we propose a novel self-supervised deep graph clustering method termed $\textbf{D}$ual $\textbf{C}$orrelation $\textbf{R}$eduction $\textbf{N}$etwork ($\textbf{DCRN}$) by reducing information correlation in a dual manner. Specifically, in our method, we first design a siamese network to encode samples. Then by forcing the cross-view sample correlation matrix and cross-view feature correlation matrix to approximate two identity matrices, respectively, we reduce the information correlation in the dual-level, thus improving the discriminative capability of the resulting features. Moreover, in order to alleviate representation collapse caused by over-smoothing in GCN, we introduce a propagation regularization term to enable the network to gain long-distance information with the shallow network structure. Extensive experimental results on six benchmark datasets demonstrate the effectiveness of the proposed DCRN against the existing state-of-the-art methods. \textit{The code of DCRN is available at \href{https://github.com/yueliu1999/DCRN}{DCRN} and a collection of deep graph clustering is shared at \href{https://github.com/yueliu1999/Awesome-Deep-Graph-Clustering}{Awesome Deep Graph Clustering} on Github}.
\end{abstract}

\section{Introduction}
Deep graph clustering is a fundamental yet challenging task whose target is to train a neural network for learning representations to divide nodes into different groups without human annotations. Thanks to the powerful graph information exploitation capability, graph convolutional networks (GCN) \cite{GCN} have recently achieved promising performance in many graph clustering applications like social networks and recommendation systems. Consequently, it has attracted considerable attention in this field and many algorithms are proposed \cite{DAEGC,ARGA,AGAE,GALA,SDCN,DFCN}.



\begin{figure}[!t]
\centering
\begin{minipage}{0.325\linewidth}
\centerline{\includegraphics[width=\textwidth]{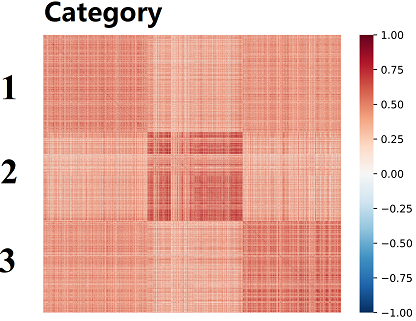}}
\vspace{3pt}
\centerline{(a) GAE}
\end{minipage}
\begin{minipage}{0.325\linewidth}
\centerline{\includegraphics[width=\textwidth]{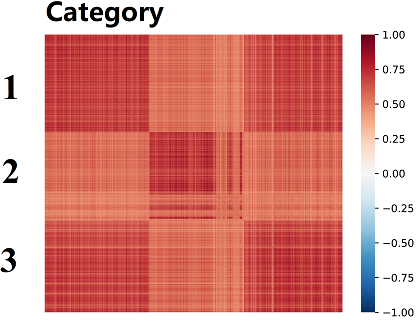}}
\vspace{3pt}
\centerline{(b) MVGRL}
\end{minipage}
\begin{minipage}{0.325\linewidth}
\centerline{\includegraphics[width=\textwidth]{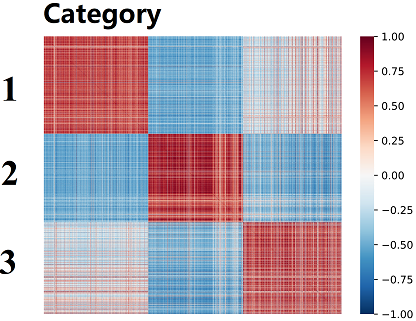}}
\vspace{3pt}
\centerline{(c) OURS}
\end{minipage}
\caption{The heat maps of node similarity matrices in the latent space of GAE \cite{GAE}, MVGRL \cite{MVGRL}, and our proposed method on the ACM dataset.
}
\label{MOTIVATION}
\end{figure}

Though good performance has been achieved, we found that the existing GCN-based clustering algorithms usually suffer from the representation collapse problem and tend to map nodes from different categories into the similar representation in the process of sample encoding. As a result, the node representation is indiscriminative and the clustering performance is limited. We illustrate this phenomenon on ACM dataset in Fig. \ref{MOTIVATION}. In this figure, we first extract the node embedding learned from three representative algorithms, i.e., the Graph Auto-Encoder (GAE) \cite{GAE}, Multi-View Graph Representation Learning (MVGRL) \cite{MVGRL}, and our proposed algorithm (OURS), and then construct the element-wise similarity matrices by calculating the cosine similarity, respectively. Finally, we visualize the similarity matrices of the three compared algorithms in Fig. \ref{MOTIVATION}. Among the compared algorithms, GAE is a classic graph convolutional network, MVGRL is a contrastive strategy enhanced algorithm, which can to some extent alleviate the representation collapse problem by introducing a positive and negative sample pair recognition mechanism.
From sub-figure (a) and (b), we observe that, in the latent space learned by both the classic algorithm and the contrastive learning enhanced algorithm, the intrinsic three dimensional cluster space is not well revealed. It indicates that representation collapse is still an open problem which is restricting the performance of GCN-based clustering algorithms.

To solve this problem, we propose a novel self-supervised deep graph clustering method termed Dual Correlation Reduction Network (DCRN) to avoid representation collapse by reducing the information correlation in a dual manner. To be specific, in our network, a dual information correlation reduction mechanism is introduced to force the cross-view sample correlation matrix and cross-view feature correlation matrix to approximate two identity matrices, respectively. In this setting, by forcing the cross-view sample-level correlation matrix to approximate an identical matrix, we guide the same noise-disturbed samples to have the identical representation while different samples to have the different representation. In this way, the sample representations would be more discriminative and in the meantime more robust against noisy information. Similarly, by letting the cross-view feature-level correlation matrix to approximate an identical matrix, the discriminative capability of latent feature is enhanced since different dimensions of the latent feature are decorrelated. This could be clearly seen in Fig. \ref{MOTIVATION} (c) since the similarity matrix generated by our proposed method can obviously exploit the hidden cluster structure among data better than the compared algorithms. As a self-supervised method, since our algorithm gets rid of the complex and space-consuming negative sample construction operations, it is more space-saving than the other contrastive learning-based algorithms. For example, in the process of model training with all samples on DBLP, CITE and ACM datasets, MVGRL spends 5753M GPU memory on average while our proposed method only spends 2672M on average. Moreover, motivated by propagation regularization \cite{p-reg}, in order to alleviate representation collapse caused by over-smoothing in GCN \cite{GCN}, we improve the long-distance information capture capability of our model with shallow network structure by introducing a propagation regularization term. This further improves the clustering performance of our proposed algorithm. The key contributions of this paper are listed as follows.

\par

\begin{itemize}
\item We propose a siamese network-based algorithm to solve the problem of representation collapse in the field of deep graph clustering.



\item A dual correlation reduction strategy is proposed to improve the discriminative capability of the sample representation. Thanks to this strategy, our method is free from the complicated negative sample generation operation and thus is more space-saving and more flexible against training batch size.

\item Extensive experimental results on six benchmark datasets demonstrate the superiority of the proposed method against the existing state-of-the-art deep graph clustering competitors.

\end{itemize}

\section{Related Work}

\subsection{Attributed Graph Clustering}
Graph Neural Networks (GNNs), which learn the representation from both node attributes and graph structures, have emerged as a powerful approach for attributed graph clustering. Specifically, GAE/VGAE \cite{GAE} embeds the node attributes with structure information via a graph encoder and then reconstructs the graph structure by an inner product decoder. Inspired by their success, recent researches, DAEGC \cite{DAEGC}, GALA \cite{GALA}, ARGA \cite{ARGA} and AGAE \cite{AGAE} further improve the early works with graph attention network, Laplacian sharpening, and generative adversarial learning. Although achieving promising clustering performance, the over-smoothing problem has not been effectively tackled in these methods, which affects the clustering performance. More recently, SDCN \cite{SDCN} and DFCN \cite{DFCN} are proposed to jointly learn an Auto-Encoder (AE) \cite{AE_K_MEANS} and a Graph Auto-Encoder (GAE) \cite{GAE} in a united framework to alleviate the over-smoothing problem via an information transport operation and a structure-attribute fusion module, respectively. Although both methods have proved that introducing the attribute features into the latent structure space can effectively address the over-smoothing issue, SDCN and DFCN suffer from another non-negligible limitation, i.e., information correlation, resulting in less discriminative representations and sub-optimal clustering performance. In contrast, our method improves the existing advanced deep graph clustering algorithm by introducing a dual information correlation reduction mechanism from the perspective of sample and feature levels to alleviate representation collapse.

\subsection{Representation Collapse}
Representation collapse, which maps all data into a same representation, is a common issue in current self-supervised representation learning methods. Some contrastive learning methods are proposed to solve this problem. MoCo \cite{MOCO} utilizes a momentum encoder to maintain the consistent representation of negative pairs drawn from a memory bank. SimCLR \cite{SIMCLR} defines the ``positive'' and ``negative'' sample pairs, and pulls closer the ``positive'' samples existing in the current batch while pushing the ``negative'' samples away. 
By replacing the empty cluster with a perturbated non-empty cluster, DeepCluster \cite{DeepCluster} is able to alleviate the collapsed representation. In addition, BYOL \cite{BYOL} and SimSiam \cite{SIMSAIM} have demonstrated that the momentum encoder and the stop-gradient mechanism are crucial to avoid representation collapse without demanding negative samples for producing prediction targets. More recently, a simple yet effective algorithm, Barlow Twins \cite{BARLOW} is proposed to alleviate the collapsed representation by reducing the redundant information between the representation of distorted samples. Inspired by its advantages, we naturally extend the idea of Barlow Twins into deep graph clustering and further design a dual correlation reduction mechanism to address representation collapse in deep clustering network. Compared to other contrastive learning methods, 
our proposed method learns the discriminative embedding to avoid collapse without negative sample generation, large batches or asymmetric mechanisms.

\section{Dual Correlation Reduction Network}
\begin{figure}[!t]
\centering
\includegraphics[scale=0.35]{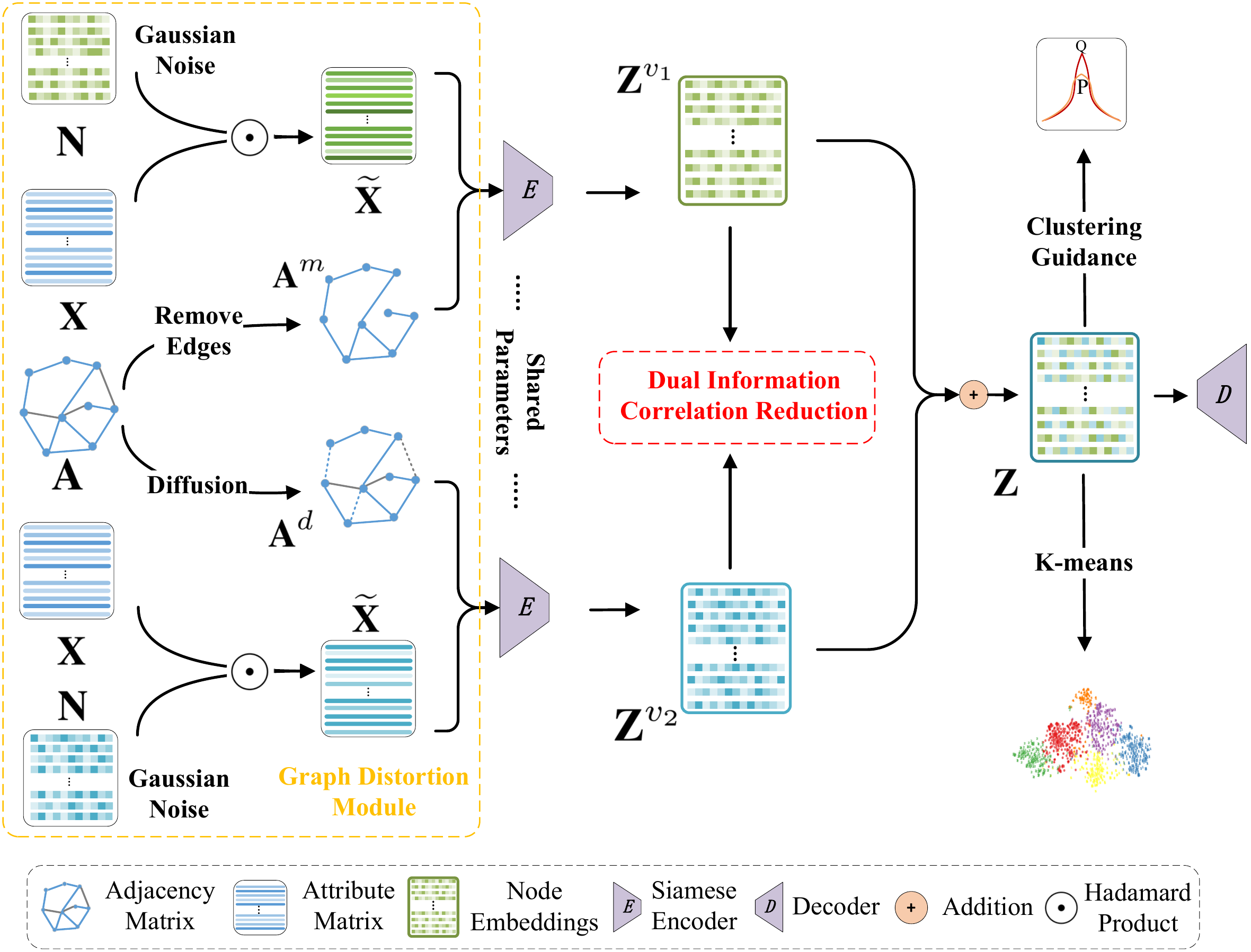} 
\caption{Illustration of the Dual Correlation Reduction Network (DCRN). In the proposed algorithm, the graph distortion module first generates two distorted graphs by introducing attribute and graph disturbances. Then, by forcing the same sample within two distorted graphs to have identical representations in both feature level and sample level, while different samples have different representations also in dual levels, the network is guided to be more discriminative with less memory consumption.
}


\label{OVERRALL_FIGURE}  
\end{figure}

We introduce a novel self-supervised deep graph clustering method termed Dual Correlation Reduction Network (DCRN), which aims to avoid representation collapse by reducing information correlation in a dual manner. As illustrated in Fig. \ref{OVERRALL_FIGURE}, DCRN mainly consists of two components, i.e., a graph distortion module and a dual information correlation reduction (DICR) module. Note that the extraction backbone network of DCRN is similar to that of DFCN \cite{DFCN}. In the following sections, We will introduce the graph distortion module, DICR module, and network objectives in detail.

\subsection{Notations and Problem Definition}

Given an undirected graph $\mathcal{G}=\left \{\mathcal{V}, \mathcal{E} \right \}$ with $C$ categories of nodes, $\mathcal{V}=\{v_1, v_2, \dots, v_N\}$ and $\mathcal{E}$ are the node set and the edge set, respectively. The graph is characterized by its attribute matrix $\textbf{X} \in \mathds{R}^{N\times D}$ and original adjacency matrix $\textbf{A}=(a_{ij})_{N\times N}$, where $a_{ij}=1$ if $(v_i,v_j)\in \mathcal{E}$, otherwise $a_{ij}=0$. The corresponding degree matrix is $\textbf{D}=diag(d_1, d_2, \dots ,d_N)\in \mathds{R}^{N\times N}$ and $d_i=\sum_{(v_i,v_j)\in \mathcal{E}}a_{ij}$. With $\textbf{D}$, the original adjacency matrix $\textbf{A}$ can be normalized as $\widetilde{\textbf{A}} \in \mathds{R}^{N\times N}$ through calculating $\textbf{D}^{-1}(\textbf{A}+\textbf{I})$, where $\textbf{I} \in \mathds{R}^{N \times N}$ is an identity matrix. In this paper, we aim to train a siamese graph encoder that embeds all nodes into the low-dimension latent space in an unsupervised manner. The resultant latent embedding can then be directly utilized to perform node clustering by K-means \cite{K-means}.
The notations are summarized in Table \ref{NOTATION_TABLE}.

\subsection{Graph Distortion Module}
Recent efforts in self-supervised graph representation learning have demonstrated that graph distortion could enable the network to learn rich representations from different contexts for nodes \cite{MVGRL, aug_2}. Inspired by their success, as illustrated in Fig. \ref{OVERRALL_FIGURE}, we consider two types of distortion on graphs, i.e., feature corruption and edge perturbation.

\noindent{\textbf{Feature Corruption}}. For the attribute-level distortion, we first sample a random noise matrix $\textbf{N} \in \mathds{R}^{N \times D}$ from a Gaussian distribution $\mathcal{N}(1, 0.1)$. Then the resulting corrupted attribute matrix $\widetilde{\textbf{X}} \in \mathds{R}^{N \times D}$ can be formulated:

\begin{equation} 
\widetilde{\textbf{X}}=\textbf{X} \odot \textbf{N},
\label{SCALING}
\end{equation}
where $\odot$ denotes the Hadamard product \cite{hadamard}.

\noindent{\textbf{Edge Perturbation}}. In addition to corrupting node features, for structure-level distortion, we introduce two strategies for edge perturbation. One is similarity-based edge removing. Thus, we first calculate the sample pair-wise cosine similarity in latent space, and then generate a masked matrix $\textbf{M} \in \mathds{R}^{N \times N}$ according to the similarity matrix, where the lowest 10\% linkage relation would be manually removed. Finally, the edge-masked adjacency matrix $\textbf{A}^{m} \in \mathds{R}^{N \times N}$ would be normalized and be computed as:

\begin{equation}
\textbf{A}^{m} = \textbf{D}^{-\frac{1}{2}}( (\textbf{A} \odot \textbf{M}) + \textbf{I} )\textbf{D}^{-\frac{1}{2}}.
\label{DROP}
\end{equation}

The other is the graph diffusion, where we follow MVGRL \cite{MVGRL} to transform the normalized adjacency matrix to a graph diffusion matrix by Personalized PageRank (PPR) \cite{PAGERANK}:

\begin{equation}
\textbf{A}^{d}=\alpha(\textbf{I}-(1-\alpha)(\textbf{D}^{-\frac{1}{2}}(\textbf{A} + \textbf{I})\textbf{D}^{-\frac{1}{2}}))^{-1},
\label{PPR}
\end{equation}
where $\alpha$ is the teleport probability that is set to 0.2. Finally, we denote $\mathcal{G}^{1}=(\widetilde{\textbf{X}},\textbf{A}^m)$ and $\mathcal{G}^{2}=(\widetilde{\textbf{X}},\textbf{A}^d)$ as two views of the graph, respectively.

\begin{table}[!t]
\centering
\small
\begin{tabular}{ll}
\toprule
Notations                                        & Meaning                                \\ \midrule
$\textbf{X}\in \mathds{R}^{N\times D}$           & Attribute matrix                       \\
$\textbf{A}\in \mathds{R}^{N\times N}$           & Original adjacency matrix              \\
$\textbf{D}\in \mathds{R}^{N\times N}$           & Degree matrix                          \\
$\widetilde{\textbf{A}}\in \mathds{R}^{N\times N}$   & Normalized adjacency matrix            \\
$\textbf{A}^m\in \mathds{R}^{N\times N}$   & 
Edge-masked adjacency matrix            \\
$\textbf{A}^d\in \mathds{R}^{N\times N}$   & 
Graph diffusion matrix            \\
$\widehat{\textbf{X}}\in \mathds{R}^{N\times D}$ & Rebuilt attribute matrix         \\
$\widehat{\textbf{A}}\in \mathds{R}^{N\times N}$ & Rebuilt adjacency matrix         \\
$\textbf{Z}^{v_k} \in \mathds{R}^{N\times d}$   & Node embedding in $k$-th view      \\
$\textbf{Z} \in \mathds{R}^{N\times d}$         & Clustering-oriented latent embedding      \\
$\widetilde{\textbf{Z}}^{v_k} \in \mathds{R}^{d\times K}$  & Cluster-level embedding in $k$-th view       \\
$\textbf{S}^\mathcal{N} \in \mathds{R}^{N\times N}$        & Cross-view sample correlation matrix  \\
$\textbf{S}^\mathcal{F} \in \mathds{R}^{d\times d}$        & Cross-view feature correlation matrix  \\
$\textbf{Q} \in \mathds{R}^{N\times C}$          & Soft assignment distribution           \\
$\textbf{P} \in \mathds{R}^{N\times C}$          & Target distribution                    \\ 
\bottomrule
\end{tabular}
\caption{Notation summary}
\label{NOTATION_TABLE} 
\end{table}

\subsection{Dual Information Correlation Reduction}
\begin{figure}[!t]
\centering
\includegraphics[scale=0.48]{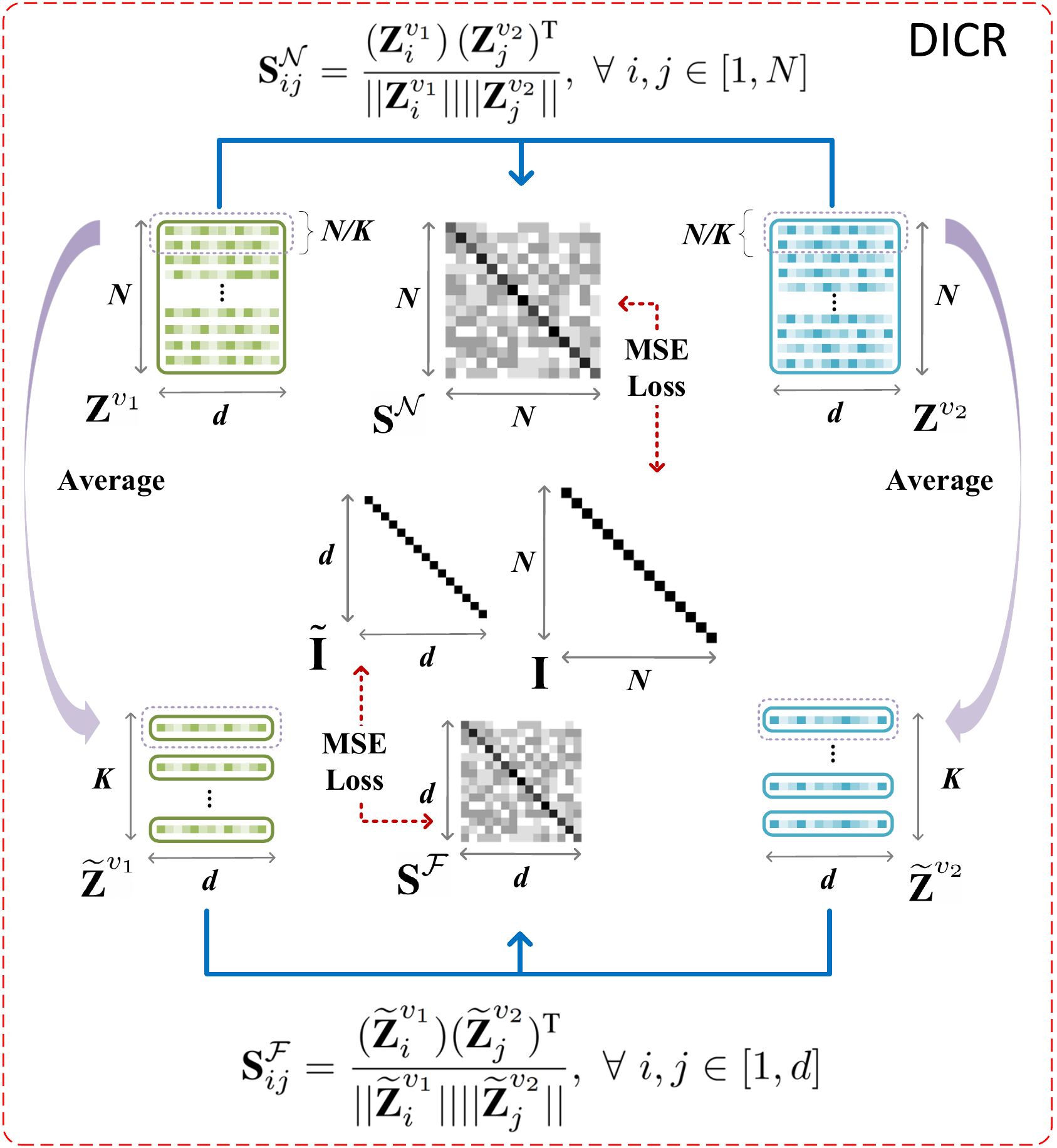} 
\caption{Illustration of Dual Information Correlation Reduction (DICR) mechanism.}
\label{DICR}  
\end{figure}

In this section, we introduce a dual information correlation reduction (DICR) mechanism to filter the redundant information of the latent embedding in a dual manner, i.e., sample-level correlation reduction (SCR) and feature-level correlation reduction (FCR), which aims to constrain our network to learn more discriminative latent features, thus alleviating representation collapse. SCR and FCR are both illustrated in Fig. \ref{DICR} in detail.


\noindent{\textbf{Sample-level Correlation Reduction}}. The learning process of SCR includes two steps. For given two-view node embeddings $\textbf{Z}^{v_1}$ and $\textbf{Z}^{v_2}$ learned by a siamese graph encoder, we firstly calculate the elements in cross-view sample correlation matrix $\textbf{S}^{\mathcal{N}} \in \mathds{R}^{N \times N}$ by: 
\begin{equation}
\textbf{S}_{ij}^{\mathcal{N}}= \frac{\left(\textbf{Z}^{v_1}_i\right) (\textbf{Z}^{v_2}_j)^{\text{T}}}{||\textbf{Z}^{v_1}_i|| || \textbf{Z}^{v_2}_j ||}, \,\,  \forall\,\,i, j \in [1, N],
\label{SAMPLE_CROSS}
\end{equation}
where $\textbf{S}_{ij}^{\mathcal{N}} \in [-1,1]$ denotes the cosine similarity between $i$-th node embedding in the first view and $j$-th node embedding in the second view. After that, we make the cross-view sample correlation matrix $\textbf{S}^{\mathcal{N}}$ to be equal to an identity matrix $\textbf{I}\in \mathds{R}^{N\times N}$, formulated as: 
\begin{equation}
\begin{aligned}
\mathcal{L}_N &= \frac{1}{N^2}\sum (\textbf{S}^{\mathcal{N}}-\textbf{I})^2 \\
 &=  \frac{1}{N}\sum\limits_{i=1}^N \left(\textbf{S}^{\mathcal{N}}_{ii}-1\right)^2
+
\frac{1}{N^2-N}\sum\limits_{i=1}^N \sum\limits_{j\ne i} \left(\textbf{S}^{\mathcal{N}}_{ij}\right)^2, 
\end{aligned}
\label{SAMPLE_LOSS}
\end{equation}
where the first term encourages the diagonal elements in $\textbf{S}^{\mathcal{N}}$ equal to 1, which indicates that the embedding of each node in two different views are enforced to agree with each other. The second term makes the off-diagonal elements in $\textbf{S}^{\mathcal{N}}$ equal to 0 to minimize the agreement between embeddings of different nodes across two views. This decorrelation operation could help our network reduce the redundant information among nodes in the latent space so that the learned embedding could be more discriminative.



\noindent{\textbf{Feature-level Correlation Reduction}}. 
Apart from building nontrivial embeddings by reducing the sample correlation across two views, we further consider to refine the information correlation from the aspect of feature dimension. Specifically, Fig. \ref{DICR} illustrates our feature-level correlation reduction design, which is implemented in three steps. First, we project two-view node embeddings $\textbf{Z}^{v_1}$ and $\textbf{Z}^{v_2}$ into cluster-level embeddings $\widetilde{\textbf{Z}}^{v_1}$ and $\widetilde{\textbf{Z}}^{v_2} \in \mathds{R}^{d \times K}$ using a readout function $\mathcal{R}(\cdot) : \mathds{R}^{d \times N} \to \mathds{R}^{d \times K}$, formulated as:
\begin{equation}
\widetilde{\textbf{Z}}^{v_k} = \mathcal{R}\left((\textbf{Z}^{v_k})^{\text{T}}\right).
\label{READOUT}
\end{equation}
Then we again calculate the cosine similarity between $\widetilde{\textbf{Z}}^{v_1}$ and $\widetilde{\textbf{Z}}^{v_2}$ along with the feature dimension, that's:
\begin{equation}
\textbf{S}_{ij}^{\mathcal{F}}= \frac{(\widetilde{\textbf{Z}}^{v_1}_i) (\widetilde{\textbf{Z}}^{v_2}_j)^\text{T}}{||\widetilde{\textbf{Z}}^{v_1}_i|| ||\widetilde{\textbf{Z}}^{v_2}_j||} , \,\,  \forall\,\,i, j \in [1, d],
\label{FEATURE_CROSS}
\end{equation}
where $\textbf{S}_{ij}^{\mathcal{F}}$ denotes the feature similarity between $i$-th dimension feature in one view and $j$-th dimension in another view. 
Thereafter, similar to the objective functions Eq. \eqref{SAMPLE_LOSS}, we make the cross-view feature correlation matrix $\textbf{S}^{\mathcal{F}}$ to be equal to an identity matrix $\widetilde{\textbf{I}} \in \mathds{R}^{d \times d}$:
\begin{equation}
\begin{aligned}
\mathcal{L}_F &= \frac{1}{{d}^2}\sum (\textbf{S}^{\mathcal{F}}-\widetilde{\textbf{I}})^2 \\ &= \frac{1}{d^2}\sum\limits_{i=1}^{d} \left(\textbf{S}^{\mathcal{F}}_{ii}-1\right)^2
+
\frac{1}{d^2-d}\sum\limits_{i=1}^{d} \sum\limits_{j\ne i} \left(\textbf{S}_{ij}^{\mathcal{F}}\right)^2,
\end{aligned}
\label{FEATURE_LOSS}
\end{equation}
where $d$ is the latent embedding dimension. Both terms in Eq. \eqref{FEATURE_LOSS} mean that the representations of the same dimension feature in two augmented views are pulled closer while others are pushed away, respectively. Finally, we combine the decorrelated latent embeddings from two views with a linear combination operation, thus the resultant clustering-oriented latent embeddings $\textbf{Z} \in \mathds{R}^{N \times d}$ can then be used to performed clustering by K-means \cite{K-means}: 
\begin{equation}
\textbf{Z} = \frac{1}{2}(\textbf{Z}^{v_1}+\textbf{Z}^{v_2}).
\label{FUSION}
\end{equation}
Technically, the proposed DICR mechanism considers the correlation reduction in both the perspective of the sample and feature level. In this way, the redundant features could be filtered while more discriminative features could be preserved in the latent space, thus the network can learn meaningful representations to avoid collapse for clustering performance improvement.


\begin{algorithm}[tb]
\small
\caption{Dual Correlation Reduction Network}
\label{ALGORITHM}
\textbf{Input}: Two-view graphs: $\mathcal{G}^{1}=(\widetilde{\textbf{X}},\textbf{A}^m)$, $\mathcal{G}^{2}=(\widetilde{\textbf{X}},\textbf{A}^d)$; Cluster number $C$; Iteration number $I$; Hyper-parameters $\gamma$ and $\lambda$. \\
\textbf{Output}: The clustering result \textbf{R}.
\begin{algorithmic}[1]

\STATE Pre-train the baseline network to obtain $\textbf{Z}$;
\STATE Initialize the cluster centers $u$ with K-means over $\textbf{Z}$;

\FOR{$i=1$ to $I$}
\STATE{Utilize the baseline network to encode $\textbf{Z}^{v_1}$ and $\textbf{Z}^{v_2}$;}
\STATE{Calculate $\widetilde{\textbf{Z}}^{v_1}$ and $\widetilde{\textbf{Z}}^{v_2}$ by Eq. \eqref{READOUT};}
\STATE Calculate $\textbf{S}^{\mathcal{N}}$ and $\textbf{S}^{\mathcal{F}}$ by Eq. \eqref{SAMPLE_CROSS} and Eq. \eqref{FEATURE_CROSS}, respectively;
\STATE Conduct the sample-level and the feature-level correlation reduction by Eq. (\ref{SAMPLE_CROSS}) and Eq. (\ref{FEATURE_CROSS}), respectively;
\STATE Fuse $\textbf{Z}^{v_1}$ and $\textbf{Z}^{v_2}$ to obtain $\textbf{Z}$ by Eq. \eqref{FUSION};

\STATE Calculate \textit{$\mathcal{L}_{DICR}$}, \textit{$\mathcal{L}_{REC}$}, and \textit{$\mathcal{L}_{KL}$}, respectively.
\STATE{Update the whole network by minimizing $\mathcal{L}$ in Eq. \eqref{LOSS}};
\ENDFOR
\STATE{Obtain \textbf{R} by performing K-means over $\textbf{Z}$.}
\STATE \textbf{return} \textbf{R}
\end{algorithmic}
\end{algorithm}

\noindent{\textbf{Propagated Regularization}}. 
Furthermore, in order to alleviate the over-smoothing phenomenon during the network training, we introduce a propagation regularization formulated as:
\begin{equation}
\mathcal{L}_{R} = JSD(\textbf{Z}, \ \widetilde{\textbf{A}}\textbf{Z}), 
\label{P-REG}
\end{equation}
where $JSD(\cdot)$ refers to the Jensen-Shannon divergence \cite{JS}. With Eq. \eqref{P-REG}, the network is able to capture long-distance information with shallow network structure to alleviate over-smoothing when the propagated information goes deeper throughout the framework. In summary, the objective of DICR module can be computed by:
\begin{equation}
\mathcal{L}_{DICR} = \mathcal{L}_N + \mathcal{L}_F + \gamma \mathcal{L}_{R}, 
\label{JOINT_DICR}
\end{equation}
where $\gamma$ is a balanced hyper-parameter.

\subsection{Objective Function}
The overall optimization objective of the proposed method consists of three parts: the loss of proposed DICR, the reconstruction loss, and the clustering loss:  
\begin{equation}
\mathcal{L} = \mathcal{L}_{DICR}+\mathcal{L}_{REC}+ \lambda \mathcal{L}_{KL}, 
\label{LOSS}
\end{equation}
where $\mathcal{L}_{REC}$ denotes the joint mean square error (MSE) reconstruction loss of node attributes and graph structure adopted in \cite{DFCN}. 
$\mathcal{L}_{KL}$ denotes the Kullback–Leibler divergence \cite{KL}, i.e., a widely-used self-supervised clustering loss \cite{DEC,IDEC,DAEGC,SDCN,DFCN}, where we generate the soft assignment distribution $\textbf{Q} \in \mathds{R}^{N \times C}$ and the target distribution $\textbf{P} \in \mathds{R}^{N \times C}$ over the clustering-oriented node embeddings $\textbf{Z}$, and then align both distributions to guide the network learning. The trade-off parameter $\lambda$ is set to 10. Here, for the design of $\mathcal{L}_{REC}$ and $\mathcal{L}_{KL}$, more details are described in the origin paper of DFCN \cite{DFCN}. The detailed learning procedure of DCRN is shown in Algorithm \ref{ALGORITHM}.

\section{Expriments}

\subsection{Datasets}
To evaluate the effectiveness of the proposed method, we conduct extensive experiments on six widely-used datasets, including DBLP, CITE, ACM\cite{SDCN}, AMAP, PUBMED, and CORAFULL\cite{AMAP}. The brief information of these datasets is summarized in Table \ref{DATASET_INFO}.

\subsection{Experiment Setup}
\subsubsection{Training Procedure}
The proposed DCRN is implemented with a NVIDIA 3090 GPU on PyTorch platform. The training process of our model includes three steps. Following DFCN \cite{DFCN}, we first pre-train the sub-networks independently with at least 30 epochs by minimizing the reconstruction loss $\mathcal{L}_{REC}$. Then both sub-networks are directly integrated into a united framework to obtain the initial clustering centers for another 100 epochs. Thereafter, we train the whole network under the guidance of Eq. (\ref{LOSS}) for 400 epochs until convergence. Finally, we perform clustering over $\textbf{Z}$ by K-means \cite{K-means}. To avoid randomness, we run each method for 10 times and report the averages with standard deviations.

\begin{table}[!t]
\centering
\small
\begin{tabular}{@{}ccccc@{}}
\toprule
Dataset  & Samples & Dimension & Edges  & Classes \\ \midrule
DBLP     & 4057    & 334       & 3528   & 4       \\
CITE     & 3327    & 3703      & 4552   & 6       \\
ACM      & 3025    & 1870      & 13128  & 3       \\
AMAP     & 7650    & 745       & 119081 & 8       \\
PUBMED   & 19717   & 500       & 44325  & 3       \\
CORAFULL & 19793   & 8710      & 63421 & 70      \\ \bottomrule
\end{tabular}
\caption{Dataset summary}
\label{DATASET_INFO} 
\end{table}

\subsubsection{Parameters Setting}
For ARGA/ARVGA \cite{ARGA}, MVGRL \cite{MVGRL}, and DFCN \cite{DFCN}, we reproduce their source code by following the setting of the original literature and present the average results. For other compared baselines, we directly report the corresponding values listed in DFCN \cite{DFCN}. For our proposed method, we adopt the code and data of DFCN for data pre-processing and testing. Besides, we adopt DFCN \cite{DFCN} as our backbone network. The network is trained with the Adam optimizer\cite{ADAM} in all experiments. The learning rate is set to 1e-3 for AMAP, 1e-4 for DBLP, 5e-5 for ACM, 1e-5 for CITE, PUBMED, and CORAFULL, respectively. The hyper-parameters $\alpha$ is set to 0.1 for PUBMED and 0.2 for other datasets. Moreover, we set $\lambda$ and $\gamma$ to 10 and 1e3, respectively. $K$ in Eq. \ref{READOUT} is set to the cluster number $C$.
\subsubsection{Metrics} 
The clustering performance is evaluated by four public metrics: Accuracy (ACC), Normalized Mutual Information (NMI), Average Rand Index (ARI) and macro F1-score (F1) \cite{LIU_1,LIU_2,LIU_3,ZHOU_1,ZHOU_2}. 
The best map between cluster ID and class ID is constructed by the Kuhn-Munkres \cite{Kuhn-Munkres}.

\begin{table*}[]
\centering
\large
\resizebox{\textwidth}{30mm}{
\begin{tabular}{@{}c|c|ccccccccccccc|c@{}}
\toprule
\textbf{Dataset}                    & \textbf{Metric} & \textbf{K-Means} & \textbf{AE} & \textbf{DEC} & \textbf{IDEC} & \textbf{GAE} & \textbf{VGAE} & \textbf{DAEGC}                    & \textbf{ARGA} & \textbf{ARVGA} & \textbf{SDCN\_Q} & \textbf{SDCN} & \textbf{MVGRL}                    & \textbf{DFCN}                     & {\color[HTML]{000000} \textbf{OURS}} \\ \midrule
                                    & ACC             & 38.65±0.65       & 51.43±0.35  & 58.16±0.56   & 60.31±0.62    & 61.21±1.22   & 58.59±0.06    & 62.05±0.48                        & 64.83±0.59    & 54.41±0.42     & 65.74±1.34       & 68.05±1.81    & 42.73±1.02                        & {\color[HTML]{3166FF} 76.00±0.80} & {\color[HTML]{FE0000} 79.66±0.25}    \\
                                    & NMI             & 11.45±0.38       & 25.40±0.16  & 29.51±0.28   & 31.17±0.50    & 30.80±0.91   & 26.92±0.06    & 32.49±0.45                        & 29.42±0.92    & 25.90±0.33     & 35.11±1.05       & 39.50±1.34    & 15.41±0.63                        & {\color[HTML]{3166FF} 43.70±1.00} & {\color[HTML]{FE0000} 48.95±0.44}    \\
                                    & ARI             & 6.97±0.39        & 12.21±0.43  & 23.92±0.39   & 25.37±0.60    & 22.02±1.40   & 17.92±0.07    & 21.03±0.52                        & 27.99±0.91    & 19.81±0.42     & 34.00±1.76       & 39.15±2.01    & 8.22±0.50                         & {\color[HTML]{3166FF} 47.00±1.50} & {\color[HTML]{FE0000} 53.60±0.46}    \\
\multirow{-4}{*}{\textbf{DBLP}}     & F1              & 31.92±0.27       & 52.53±0.36  & 59.38±0.51   & 61.33±0.56    & 61.41±2.23   & 58.69±0.07    & 61.75±0.67                        & 64.97±0.66    & 55.37±0.40     & 65.78±1.22       & 67.71±1.51    & 40.52±1.51                        & {\color[HTML]{3166FF} 75.70±0.80} & {\color[HTML]{FE0000} 79.28±0.26}    \\ \midrule
                                    & ACC             & 39.32±3.17       & 57.08±0.13  & 55.89±0.20   & 60.49±1.42    & 61.35±0.80   & 60.97±0.36    & 64.54±1.39                        & 61.07±0.49    & 59.31±1.38     & 61.67±1.05       & 65.96±0.31    & 68.66±0.36                        & {\color[HTML]{3166FF} 69.50±0.20} & {\color[HTML]{FE0000} 70.86±0.18}    \\
                                    & NMI             & 16.94±3.22       & 27.64±0.08  & 28.34±0.30   & 27.17±2.40    & 34.63±0.65   & 32.69±0.27    & 36.41±0.86                        & 34.40±0.71    & 31.80±0.81     & 34.39±1.22       & 38.71±0.32    & 43.66±0.40                        & {\color[HTML]{3166FF} 43.90±0.20} & {\color[HTML]{FE0000} 45.86±0.35}    \\
                                    & ARI             & 13.43±3.02       & 29.31±0.14  & 28.12±0.36   & 25.70±2.65    & 33.55±1.18   & 33.13±0.53    & 37.78±1.24                        & 34.32±0.70    & 31.28±1.22     & 35.50±1.49       & 40.17±0.43    & 44.27±0.73                        & {\color[HTML]{3166FF} 45.50±0.30} & {\color[HTML]{FE0000} 47.64±0.30}    \\
\multirow{-4}{*}{\textbf{CITE}}     & F1              & 36.08±3.53       & 53.80±0.11  & 52.62±0.17   & 61.62±1.39    & 57.36±0.82   & 57.70±0.49    & 62.20±1.32                        & 58.23±0.31    & 56.05±1.13     & 57.82±0.98       & 63.62±0.24    & 63.71±0.39                        & {\color[HTML]{3166FF} 64.30±0.20} & {\color[HTML]{FE0000} 65.83±0.21}    \\ \midrule
                                    & ACC             & 67.31±0.71       & 81.83±0.08  & 84.33±0.76   & 85.12±0.52    & 84.52±1.44   & 84.13±0.22    & 86.94±2.83                        & 86.29±0.36    & 83.89±0.54     & 86.95±0.08       & 90.45±0.18    & 86.73±0.76                        & {\color[HTML]{3166FF} 90.90±0.20} & {\color[HTML]{FE0000} 91.93±0.20}    \\
                                    & NMI             & 32.44±0.46       & 49.30±0.16  & 54.54±1.51   & 56.61±1.16    & 55.38±1.92   & 53.20±0.52    & 56.18±4.15                        & 56.21±0.82    & 51.88±1.04     & 58.90±0.17       & 68.31±0.25    & 60.87±1.40                        & {\color[HTML]{3166FF} 69.40±0.40} & {\color[HTML]{FE0000} 71.56±0.61}    \\
                                    & ARI             & 30.60±0.69        & 54.64±0.16  & 60.64±1.87   & 62.16±1.50    & 59.46±3.10   & 57.72±0.67    & 59.35±3.89                        & 63.37±0.86    & 57.77±1.17     & 65.25±0.19       & 73.91±0.40    & 65.07±1.76                        & {\color[HTML]{3166FF} 74.90±0.40} & {\color[HTML]{FE0000} 77.56±0.52}    \\
\multirow{-4}{*}{\textbf{ACM}}      & F1              & 67.57±0.74       & 82.01±0.08  & 84.51±0.74   & 85.11±0.48    & 84.65±1.33   & 84.17±0.23    & 87.07±2.79                        & 86.31±0.35    & 83.87±0.55     & 86.84±0.09       & 90.42±0.19    & 86.85±0.72                        & {\color[HTML]{3166FF} 90.80±0.20} & {\color[HTML]{FE0000} 91.94±0.20}    \\ \midrule
                                    & ACC             & 27.22±0.76       & 48.25±0.08  & 47.22±0.08   & 47.62±0.08    & 71.57±2.48   & 74.26±3.63    & 76.44±0.01                        & 69.28±2.30    & 61.46±2.71     & 35.53±0.39       & 53.44±0.81    & 45.19±1.79                        & {\color[HTML]{3166FF} 76.88±0.80} & {\color[HTML]{FE0000} 79.94±0.13}    \\
                                    & NMI             & 13.23±1.33       & 38.76±0.30  & 37.35±0.05   & 37.83±0.08    & 62.13±2.79   & 66.01±3.40    & 65.57±0.03                        & 58.36±2.76    & 53.25±1.91     & 27.90±0.40       & 44.85±0.83    & 36.89±1.31                        & {\color[HTML]{3166FF} 69.21±1.00} & {\color[HTML]{FE0000} 73.70±0.24}    \\
                                    & ARI             & 5.50±0.44        & 20.80±0.47  & 18.59±0.04   & 19.24±0.07    & 48.82±4.57   & 56.24±4.66    & {\color[HTML]{3166FF} 59.39±0.02} & 44.18±4.41    & 38.44±4.69     & 15.27±0.37       & 31.21±1.23    & 18.79±0.47                        & 58.98±0.84                        & {\color[HTML]{FE0000} 63.69±0.20}    \\
\multirow{-4}{*}{\textbf{AMAP}}     & F1              & 23.96±0.51       & 47.87±0.20  & 46.71±0.12   & 47.20±0.11    & 68.08±1.76   & 70.38±2.98    & 69.97±0.02                        & 64.30±1.95    & 58.50±1.70      & 34.25±0.44       & 50.66±1.49    & 39.65±2.39                        & {\color[HTML]{3166FF} 71.58±0.31} & {\color[HTML]{FE0000} 73.82±0.12}    \\ \midrule
                                    & ACC             & 59.83±0.01       & 63.07±0.31  & 60.14±0.09   & 60.70±0.34    & 62.09±0.81   & 68.48±0.77    & 68.73±0.03                        & 65.26±0.12    & 64.25±1.24     & 64.39±0.30       & 64.20±1.30    & 67.01±0.52                        & {\color[HTML]{3166FF} 68.89±0.07} & {\color[HTML]{FE0000} 69.87±0.07}    \\
                                    & NMI             & 31.05±0.02       & 26.32±0.57  & 22.44±0.14   & 23.67±0.29    & 23.84±3.54   & 30.61±1.71    & 28.26±0.03                        & 24.80±0.17    & 23.88±1.05     & 26.67±1.31       & 22.87±2.04    & {\color[HTML]{3166FF} 31.59±1.45} & 31.43±0.13                        & {\color[HTML]{FE0000} 32.20±0.08}    \\
                                    & ARI             & 28.10±0.01       & 23.86±0.67  & 19.55±0.13   & 20.58±0.39    & 20.62±1.39   & 30.15±1.23    & 29.84±0.04                        & 24.35±0.17    & 22.82±1.52     & 24.61±1.46       & 22.30±2.07    & 29.42±1.06                        & {\color[HTML]{3166FF} 30.64±0.11} & {\color[HTML]{FE0000} 31.41±0.12}    \\
\multirow{-4}{*}{\textbf{PUBMED}}   & F1              & 58.88±0.01       & 64.01±0.29  & 61.49±0.10   & 62.41±0.32    & 61.37±0.85   & 67.68±0.89    & {\color[HTML]{3166FF} 68.23±0.02} & 65.69±0.13    & 64.51±1.32     & 65.46±0.39       & 65.01±1.21    & 67.07±0.36                        & 68.10±0.07                        & {\color[HTML]{FE0000} 68.94±0.08}    \\ \midrule
                                    & ACC             & 26.27±1.10       & 33.12±0.19  & 31.92±0.45   & 32.19±0.31    & 29.60±0.81   & 32.66±1.29    & 34.35±1.00                        & 22.07±0.43    & 29.57±0.59     & 29.75±0.69       & 26.67±0.40    & 31.52±2.95                        & {\color[HTML]{3166FF} 37.51±0.81} & {\color[HTML]{FE0000} 38.80±0.60}    \\
                                    & NMI             & 34.68±0.84       & 41.53±0.25  & 41.67±0.24   & 41.64±0.28    & 45.82±0.75   & 47.38±1.59    & 49.16±0.73                        & 41.28±0.25    & 48.77±0.44     & 40.10±0.22       & 37.38±0.39    & 48.99±3.95                        & {\color[HTML]{3166FF} 51.30±0.41} & {\color[HTML]{FE0000} 51.91±0.35}    \\
                                    & ARI             & 9.35±0.57        & 18.13±0.27  & 16.98±0.29   & 17.17±0.22    & 17.84±0.86   & 20.01±1.38    & 22.60±0.47                        & 12.38±0.24    & 18.80±0.57     & 16.47±0.38       & 13.63±0.27    & 19.11±2.63                        & {\color[HTML]{3166FF} 24.46±0.48} & {\color[HTML]{FE0000} 25.25±0.49}    \\
\multirow{-4}{*}{\textbf{CORAFULL}} & F1              & 22.57±1.09       & 28.40±0.30   & 27.71±0.58   & 27.72±0.41    & 25.95±0.75   & 29.06±1.15    & 26.96±1.33                        & 18.85±0.41    & 25.43±0.62     & 24.62±0.53       & 22.14±0.43    & 26.51±2.87                        & {\color[HTML]{3166FF} 31.22±0.87} & {\color[HTML]{FE0000} 31.68±0.76}    \\ \bottomrule
\end{tabular}
}
\caption{The average clustering performance with mean±std on six benchmarks. The {\color[HTML]{FE0000} red} and {\color[HTML]{3166FF} blue} values indicate the best and the runner-up results, respectively.}
\label{COMPARE_RESULT}
\end{table*}

\subsection{Performance Comparison}
To demonstrate the superiority of the proposed method, we adopt 13  baselines for performance comparisons. Specifically, 
K-means \cite{K-means} is one of the most classic traditional clustering methods. Three representative deep generative methods, i.e., AE \cite{AE_K_MEANS}, DEC \cite{DEC}, and IDEC \cite{IDEC}, train an auto-encoder and then perform a clustering algorithm over the learned latent embedding. GAE/VGAE \cite{GAE}, DAEGC \cite{DAEGC}, and ARGA/ARVGA \cite{ARGA} are three typical GCN-based frameworks that learn the representation for clustering by considering both node attribute and structure information. Furthermore, we report the performance of three state-of-the-art deep clustering methods, i.e., SDCN/SDCN$_{Q}$ \cite{SDCN}, DFCN \cite{DFCN}, and MVGRL \cite{MVGRL}, which utilize two sub-networks to process augmented graphs independently.

\begin{figure*}[!t]
\tiny
\begin{minipage}{0.139\linewidth}
\centerline{\includegraphics[width=\textwidth]{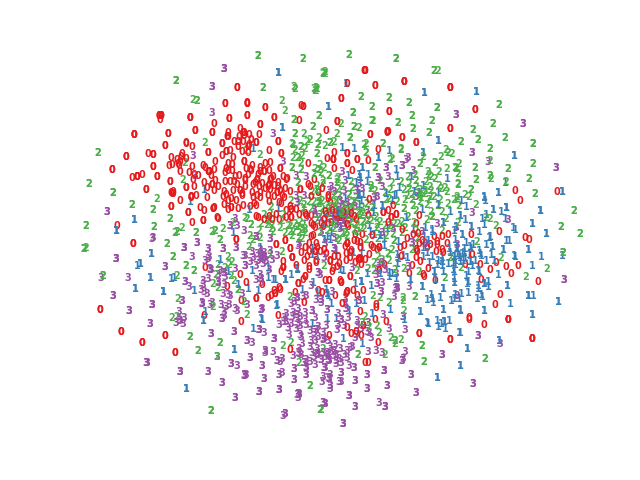}}
\centerline{\includegraphics[width=\textwidth]{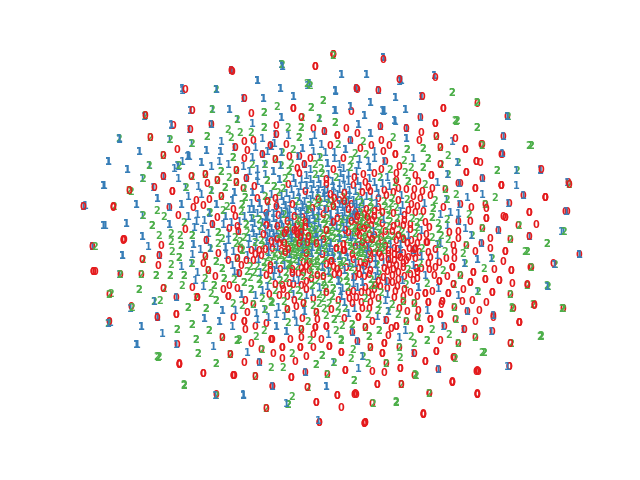}}
\centerline{Raw Data}
\end{minipage}
\begin{minipage}{0.139\linewidth}
\centerline{\includegraphics[width=\textwidth]{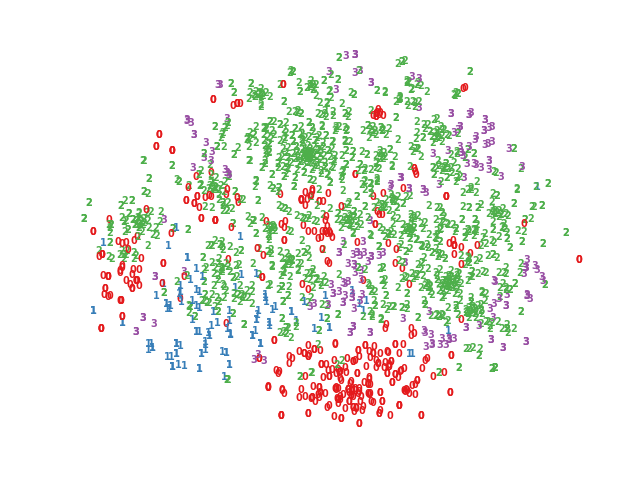}}
\centerline{\includegraphics[width=\textwidth]{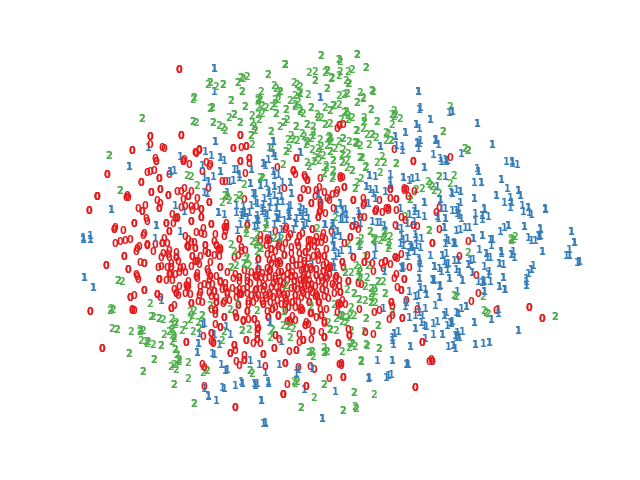}}
\centerline{AE}
\end{minipage}
\begin{minipage}{0.139\linewidth}
\centerline{\includegraphics[width=\textwidth]{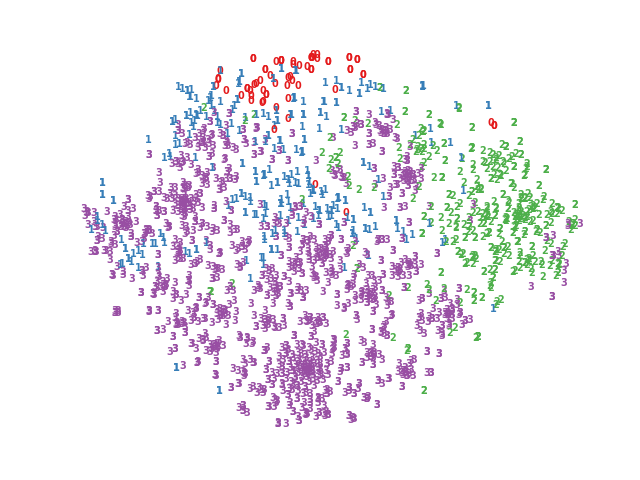}}
\centerline{\includegraphics[width=\textwidth]{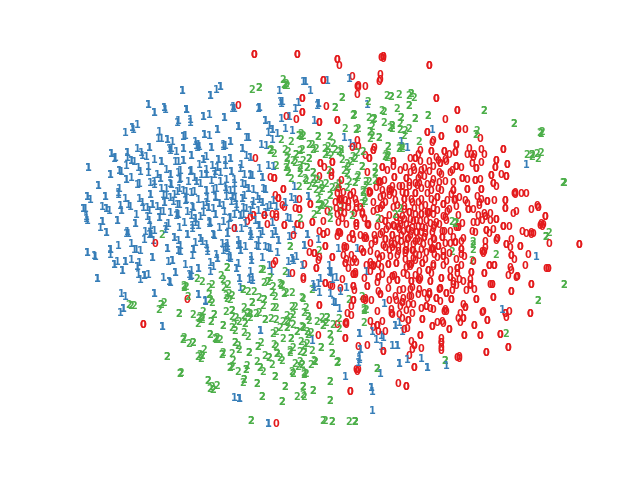}}
\centerline{DEC}
\end{minipage}
\begin{minipage}{0.139\linewidth}
\centerline{\includegraphics[width=\textwidth]{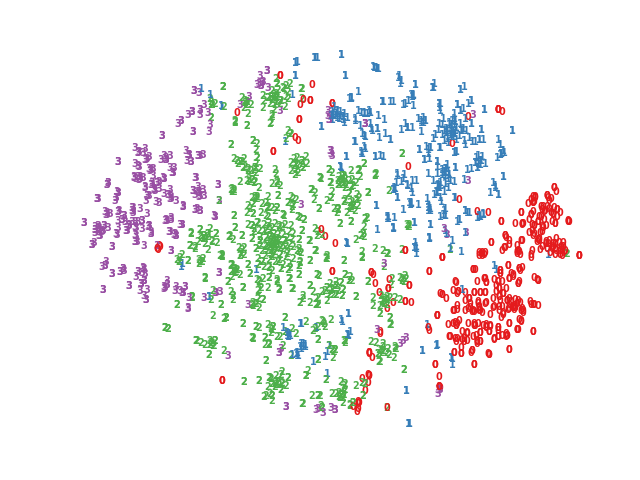}}
\centerline{\includegraphics[width=\textwidth]{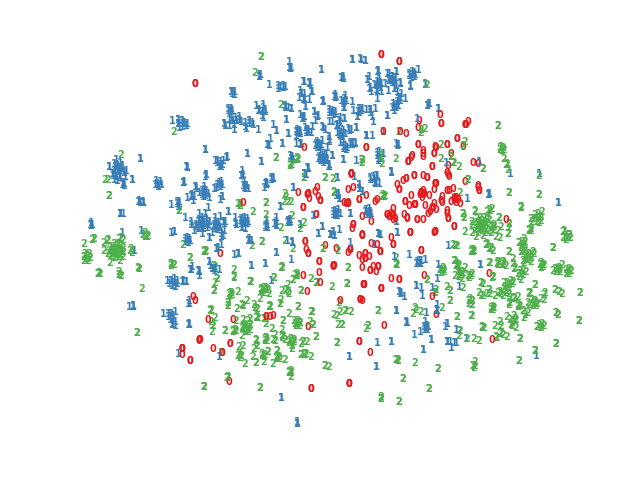}}
\centerline{GAE}
\end{minipage}
\begin{minipage}{0.139\linewidth}
\centerline{\includegraphics[width=\textwidth]{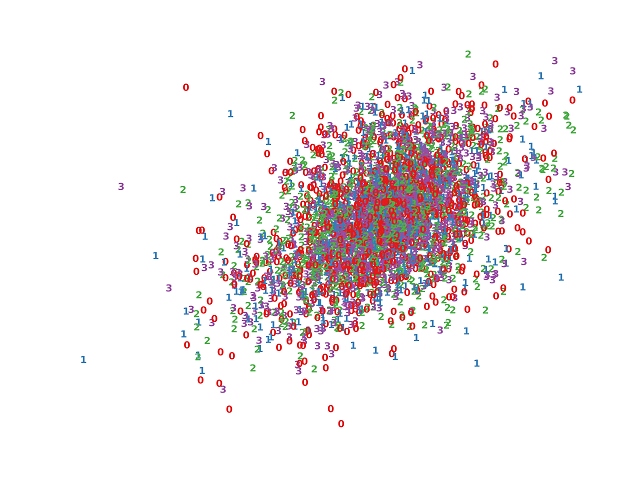}}
\centerline{\includegraphics[width=\textwidth]{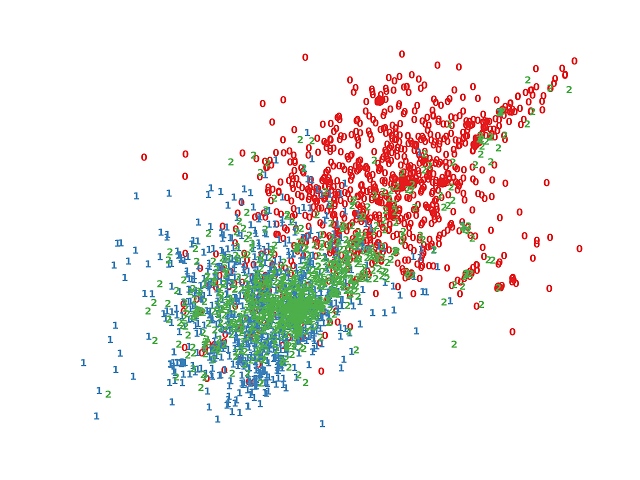}}
\centerline{ARGA}
\end{minipage}
\begin{minipage}{0.139\linewidth}
\centerline{\includegraphics[width=\textwidth]{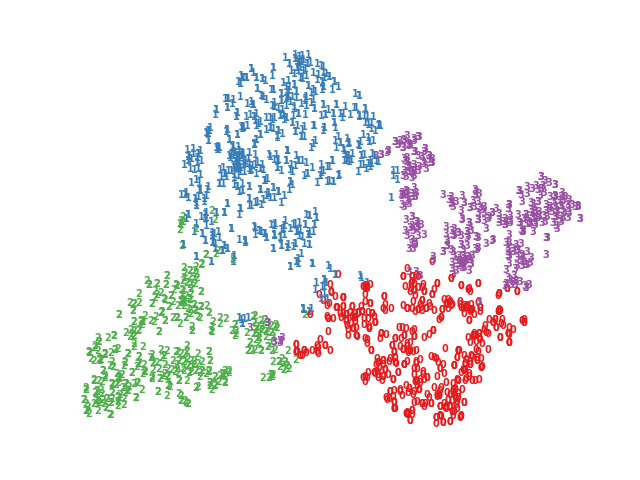}}
\centerline{\includegraphics[width=\textwidth]{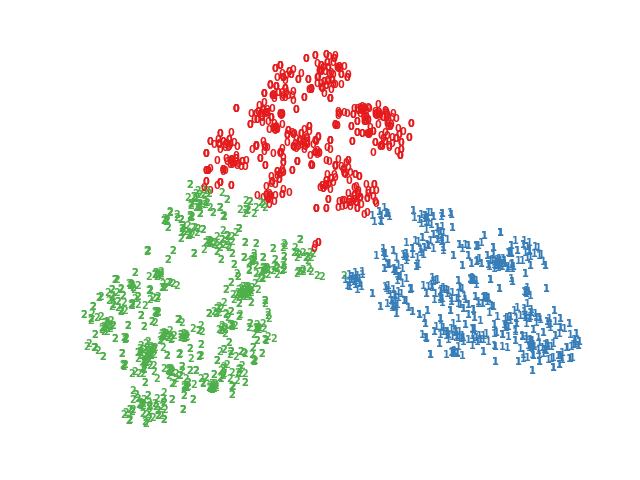}}
\centerline{DFCN}
\end{minipage}
\begin{minipage}{0.139\linewidth}
\centerline{\includegraphics[width=\textwidth]{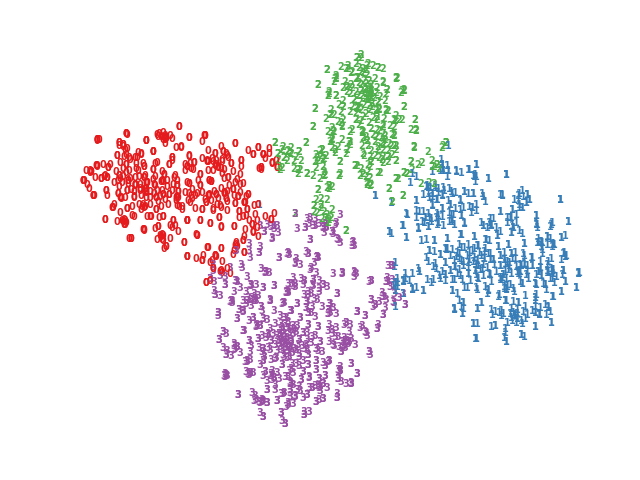}}
\centerline{\includegraphics[width=\textwidth]{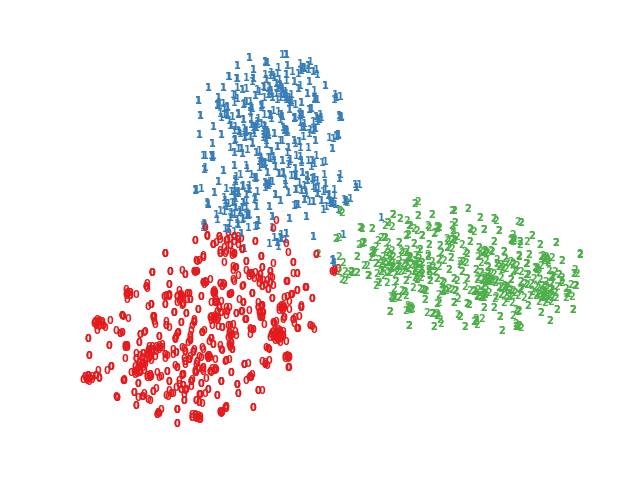}}
\centerline{OURS}
\end{minipage}

\caption{2D visualization on two datasets. The first row and second row correspond to DBLP and ACM, respectively.}
\label{VIS}  
\end{figure*}

Table \ref{COMPARE_RESULT} reports the clustering performance of all compared methods on six benchmarks. From these results, we can conclude that 1) DCRN consistently outperforms all compared methods in terms of four metrics over all datasets. SDCN/$\text{SDCN}_Q$ \cite{SDCN}, MVGRL \cite{MVGRL} and DFCN \cite{DFCN} have been considered as three strongest deep clustering frameworks. Taking the results on DBLP for example, our DCRN exceeds DFCN by 3.66\% 5.25\%, 6.60\% 3.58\% increments with respect to ACC, NMI, ARI and F1. This is because both SDCN and DFCN overly introduce the attribute information learned by the auto-encoder part into the latent space, so that the node embedding contains redundant attributes about the sample, leading to representation collapse. In contrast, by reducing the information correlation in a dual manner, DCRN can learn more meaningful representation to improve the clustering performance; 2) it can be observed that the GCN-based clustering methods GAE/VGAE \cite{GAE}, ARGA \cite{ARGA} and DAEGC \cite{DAEGC} are not comparable with ours. This is because these methods do not consider to handle information correlation redundancy, thus resulting in the trivial constant representation; 3) our method improves the auto-encoder-based clustering methods, i.e., AE \cite{AE_K_MEANS}, DEC \cite{AE_K_MEANS} and IDEC \cite{IDEC}, by a large margin, all of which have been verified strong representation learning capacity for clustering on non-graph data, while these methods that merely rely on attribute information can not effectively learn discriminative information on graphs; 4) since K-means \cite{K-means} is directly performed on raw attributes, thus achieving unpromising results.  
Overall, the aforementioned observations have demonstrated the effectiveness of our proposed method in solving representation collapse issue. In the following section, ablation studies of each module in DCRN will be introduced in detail.

\begin{table}[!t]
\centering
\tiny
\begin{tabular}{@{}c|c|cccc@{}}
\toprule
Dataset                   & Metric & Baseline   & Baseline-P & Baseline-D & Baseline-P-D \\ \midrule
\multirow{4}{*}{DBLP}     & ACC    & 76.00±0.80 & 77.00±0.41    & 79.63±0.27 & 79.66±0.25   \\
                          & NMI    & 43.70±1.00 & 44.98±0.56 & 48.95±0.48 & 48.95±0.44   \\
                          & ARI    & 47.00±1.50 & 48.51±0.84 & 53.48±0.51 & 53.60±0.46    \\
                          & F1     & 75.70±0.80  & 76.77±0.38 & 79.26±0.28 & 79.28±0.26   \\ \midrule
\multirow{4}{*}{CITE}     & ACC    & 69.50±0.20  & 70.07±0.21 & 70.88±0.19 & 70.86±0.18   \\
                          & NMI    & 43.90±0.20 & 44.75±0.40  & 45.92±0.35 & 45.86±0.35   \\
                          & ARI    & 45.50±0.30 & 46.52±0.36 & 47.73±0.29 & 47.64±0.30    \\
                          & F1     & 64.30±0.20  & 65.03±0.23 & 65.79±0.20  & 65.83±0.21   \\ \midrule
\multirow{4}{*}{ACM}      & ACC    & 90.90±0.20    & 91.57±0.12 & 91.91±0.21 & 91.93±0.20    \\
                          & NMI    & 69.40±0.40 & 70.82±0.25 & 71.56±0.61 & 71.56±0.52   \\
                          & ARI    & 74.90±0.40  & 76.68±0.28 & 77.50±0.53 & 77.56±0.52   \\
                          & F1     & 90.80±0.20 & 91.53±0.12 & 91.90±0.21  & 91.94±0.20   \\ \midrule
\multirow{4}{*}{AMAP}     & ACC    & 76.88±0.80 & 79.01±0.01 & 79.95±0.04 & 79.94±0.13   \\
                          & NMI    & 69.21±1.00  & 72.29±0.01 & 73.69±0.05 & 73.70±0.24    \\
                          & ARI    & 58.98±0.84 & 62.1±0.01  & 63.70±0.05  & 63.69±0.20   \\
                          & F1     & 71.58±0.31 & 73.09±0.00    & 73.84±0.03 & 73.82±0.12   \\ \midrule
\multirow{4}{*}{PUBMED}   & ACC    & 68.89±0.07 & 69.43±0.05 & 69.74±0.06 & 69.87±0.07   \\
                          & NMI    & 31.43±0.13 & 31.98±0.12 & 32.04±0.06 & 32.20±0.08    \\
                          & ARI    & 30.64±0.11 & 31.35±0.12 & 31.14±0.11 & 31.41±0.12   \\
                          & F1     & 68.10±0.07  & 68.54±0.06 & 68.81±0.07 & 68.94±0.08   \\ \midrule
\multirow{4}{*}{CORAFULL} & ACC    & 37.51±0.81 & 37.04±0.71 & 38.23±0.59 & 38.80±0.60    \\
                          & NMI    & 51.30±0.41  & 51.90±0.26  & 50.85±0.36 & 51.91±0.35   \\
                          & ARI    & 24.46±0.48 & 24.13±0.51 & 24.83±0.37 & 25.25±0.49   \\
                          & F1     & 31.22±0.87 & 30.35±0.87 & 31.34±0.81 & 31.68±0.76   \\ \bottomrule
                        
\end{tabular}

\caption{Ablation comparisons of DICR mechanism and the propagated regularization on six datasets.}
\label{ICR_ABLATION} 
\end{table}

\subsection{Ablation Studies}
\subsubsection{Effectiveness of DICR Module}
We conduct an ablation study to clearly verify the effectiveness of DICR module and report the results in Table \ref{ICR_ABLATION}. Here we denote the DFCN \cite{DFCN} as the Baseline since it’s the feature extraction backbone of our network. Baseline-P, Baseline-D, and Baseline-P-D denote that the baseline adopts the propagated regularization, the DICR mechanism, and both. From the results in Table \ref{ICR_ABLATION}, we can observe that 1) compare with the baseline, Baseline-P has about 0.5\% to 1.0\% performance improvement in terms of four metrics on DBLP dataset. These results demonstrate that introducing a regularization term into the network training could improve the generalization capacity of the model as well as alleviate the over-smoothing; 2) Baseline-D consistently achieves better performance than that of the baseline. Taking the results on DBLP for example, Baseline-D exceeds the baseline by 3.63\%, 5.25\%, 6.48\%, 3.56\% performance increment with respect to ACC, NMI, ARI and F1. It benefits from that we conduct a DICR mechanism to enhance the discriminative capacity of the latent embedding for clustering performance improvement. We can obtain similar conclusions from the results on other datasets; 3) the results in the last column of Table \ref{ICR_ABLATION} further verify the effectiveness of both components. As seen, Baseline-P-D achieves the best results compared to other variants.

\subsubsection{Effectiveness of Dual Level Correlation Reduction}
\begin{figure}[!t]
\centering
\tiny
\begin{minipage}{0.48\linewidth}
\centerline{\includegraphics[width=0.9\textwidth]{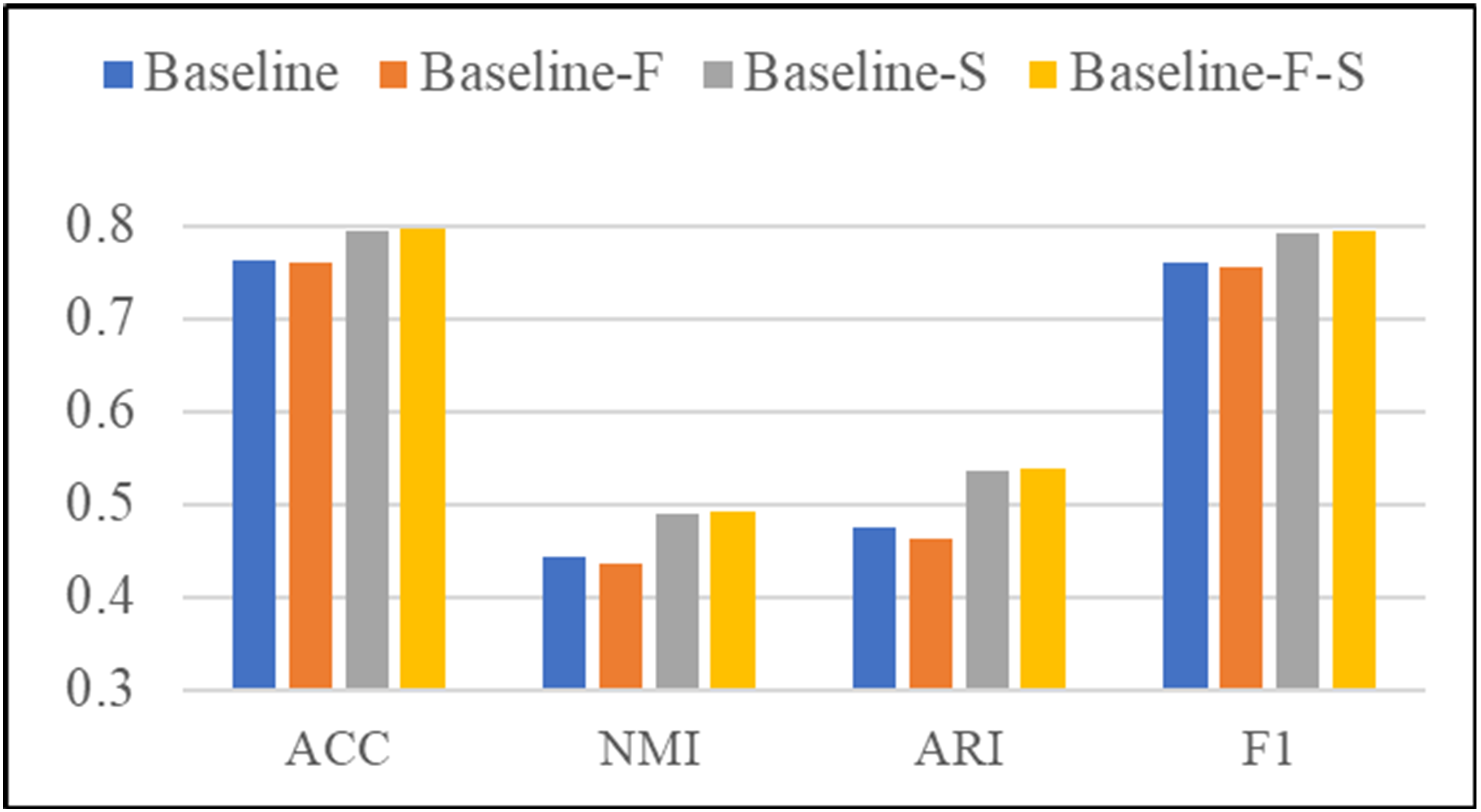}}
\vspace{3pt}
\centerline{DBLP}
\vspace{3pt}
\centerline{\includegraphics[width=0.9\textwidth]{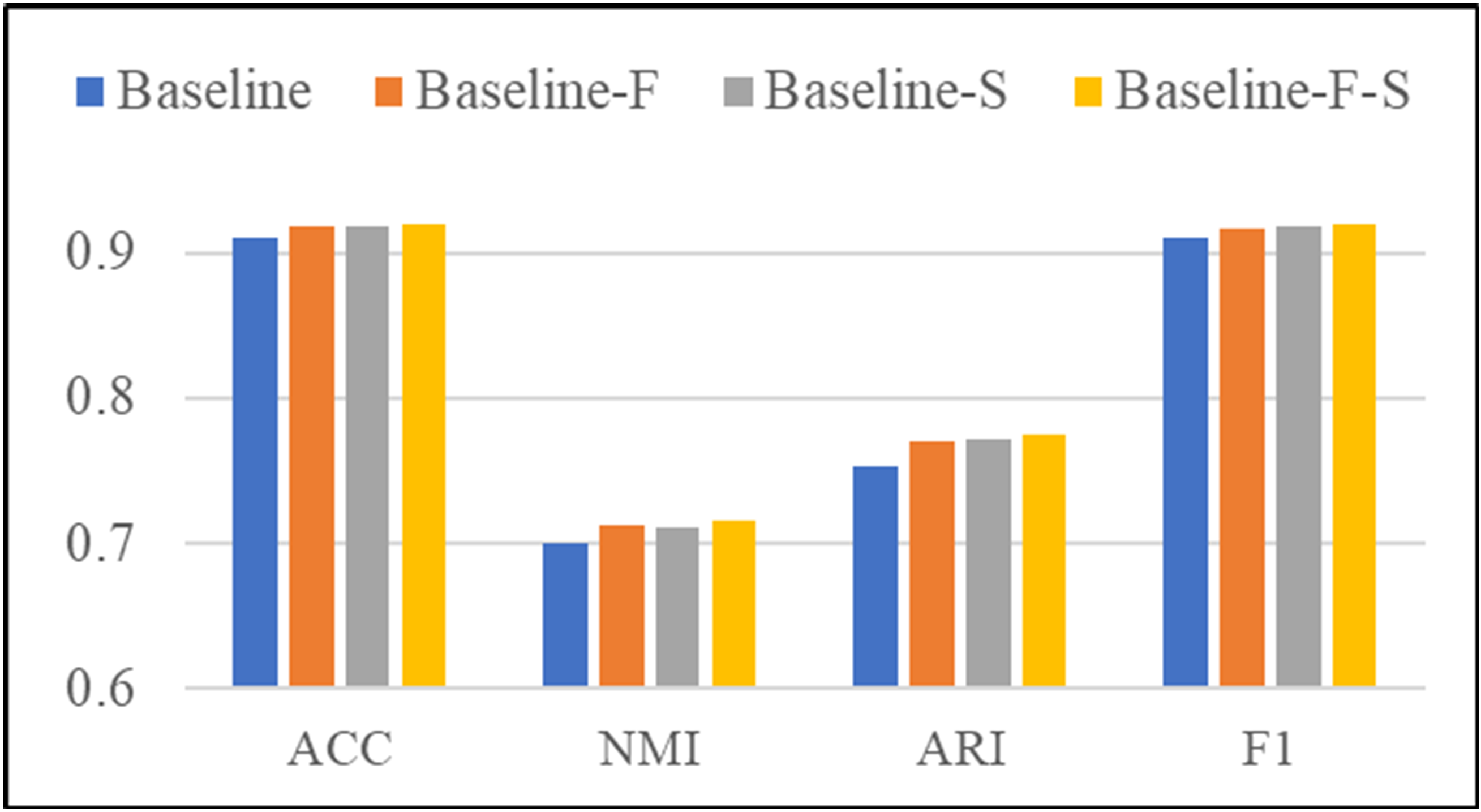}}
\vspace{3pt}
\centerline{ACM}
\vspace{3pt}
\centerline{\includegraphics[width=0.9\textwidth]{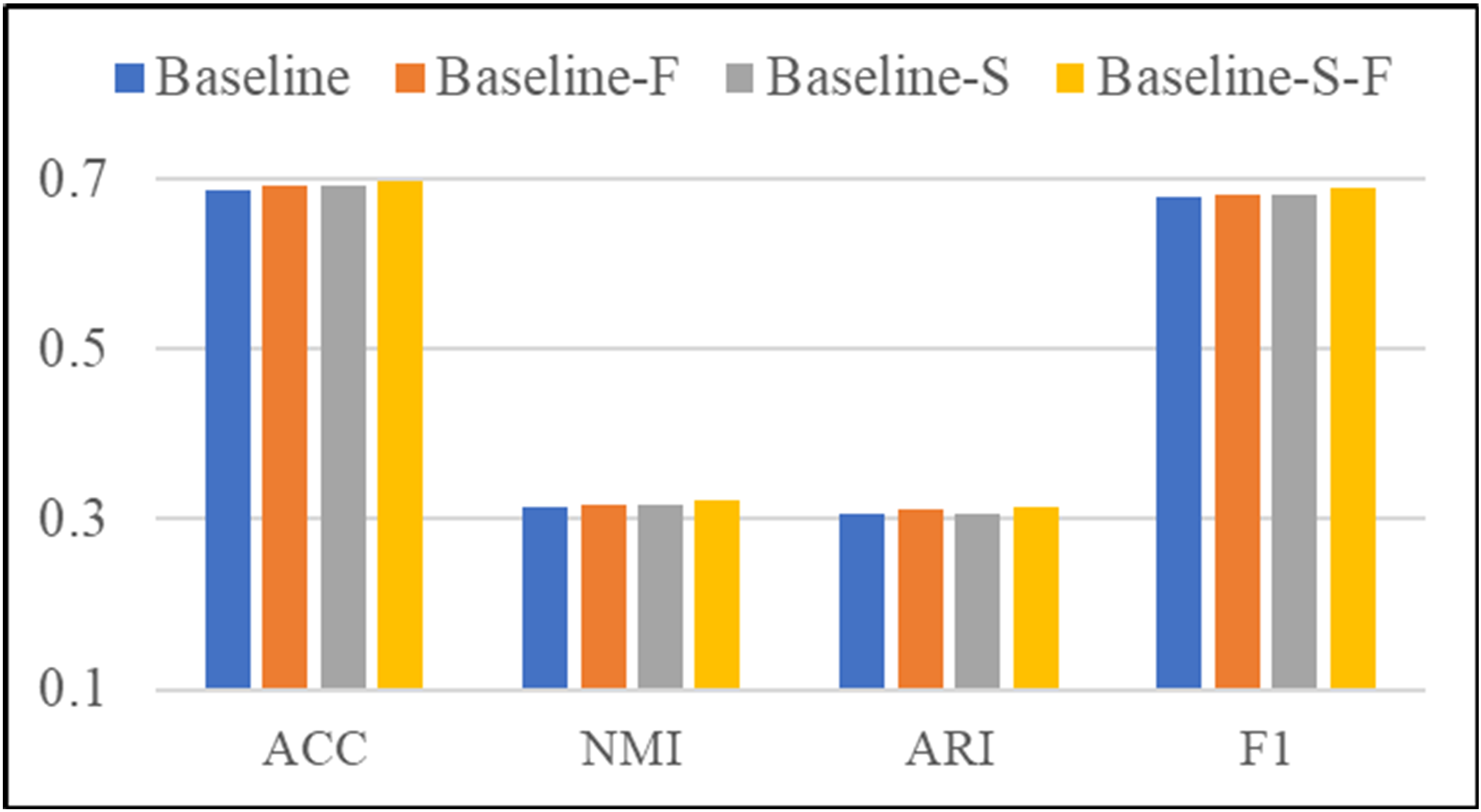}}
\vspace{3pt}
\centerline{PUBMED}

\end{minipage}
\begin{minipage}{0.48\linewidth}
\centerline{\includegraphics[width=0.9\textwidth]{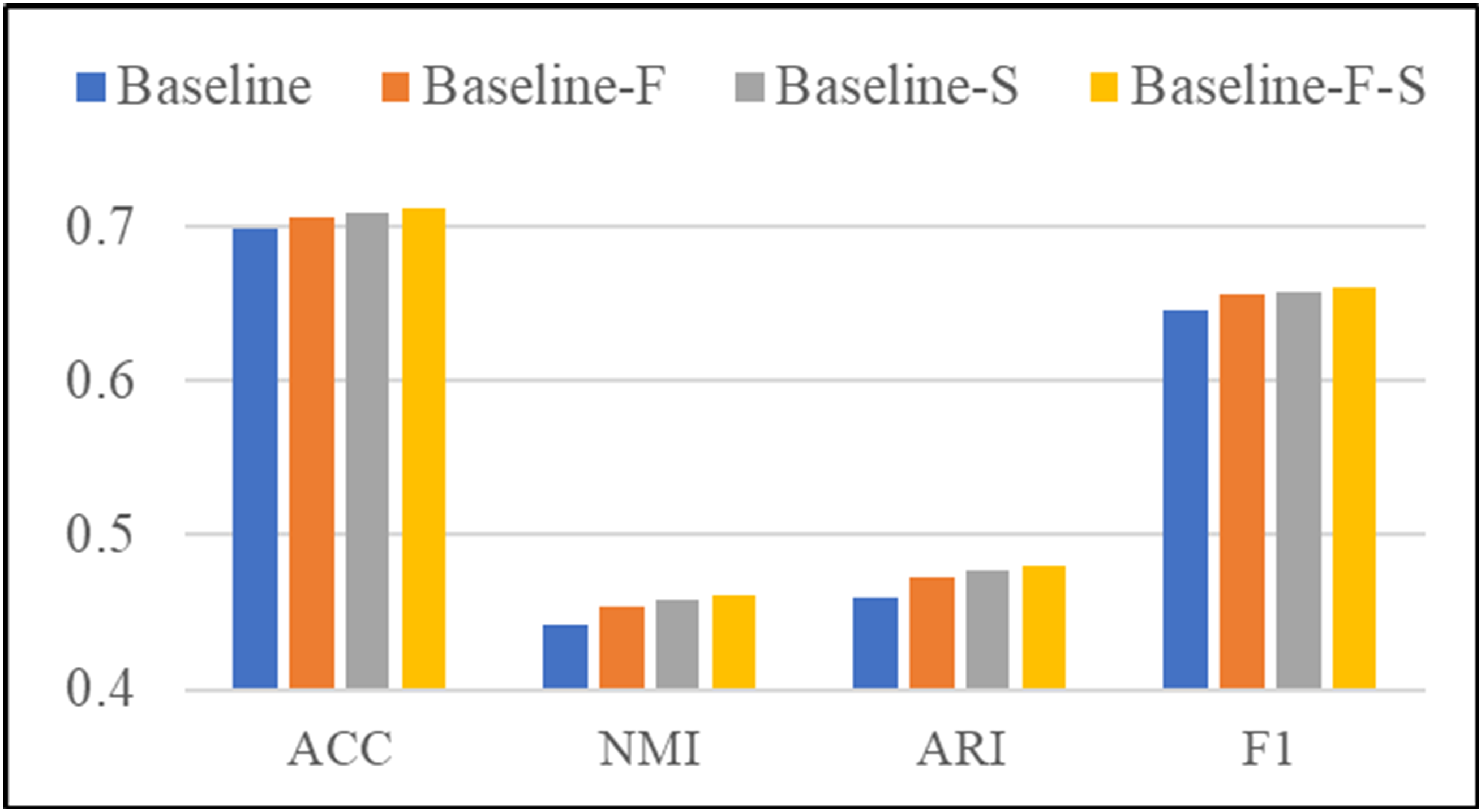}}
\vspace{3pt}
\centerline{CITE}
\vspace{3pt}
\centerline{\includegraphics[width=0.9\textwidth]{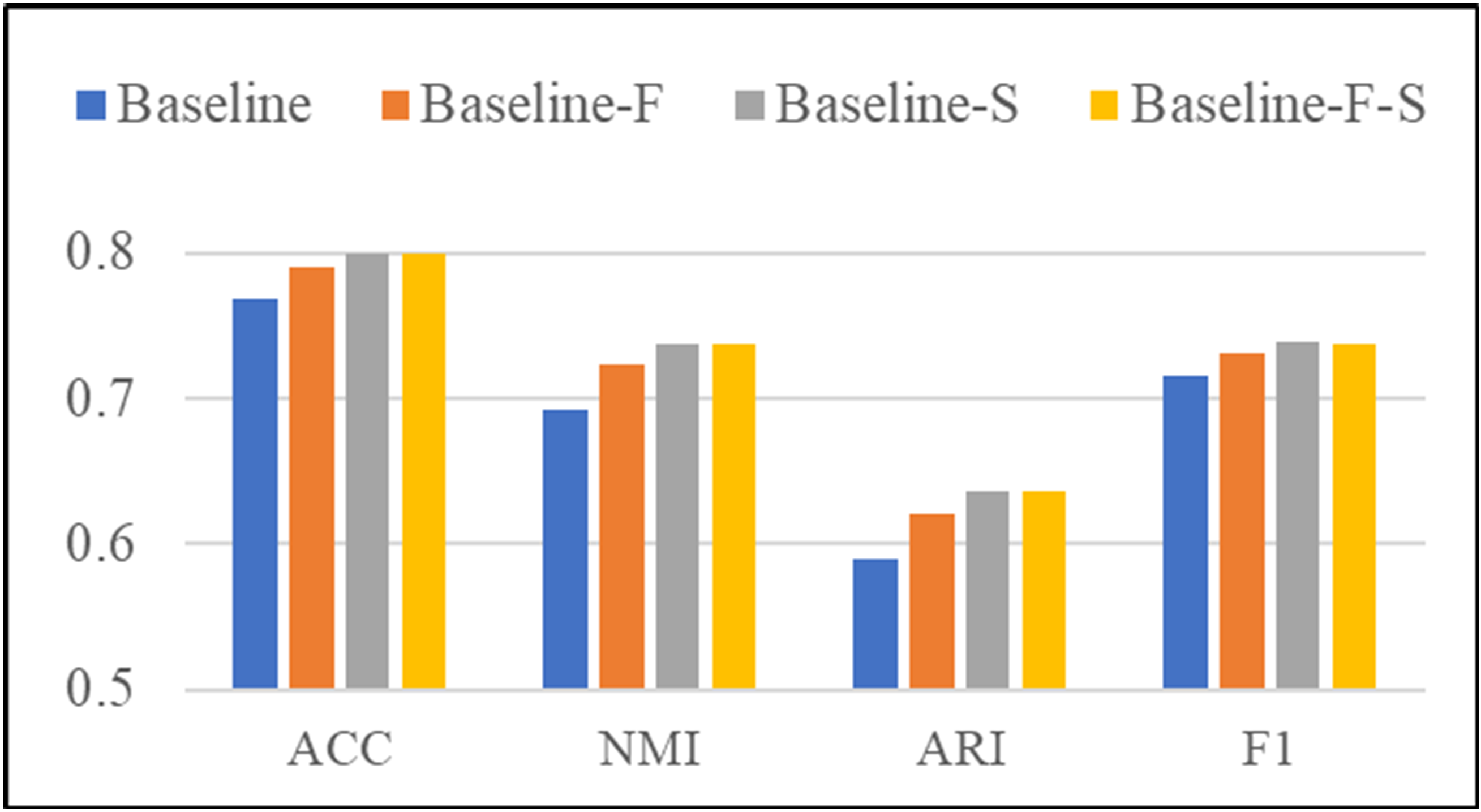}}
\vspace{3pt}
\centerline{AMAP}
\vspace{3pt}
\centerline{\includegraphics[width=0.9\textwidth]{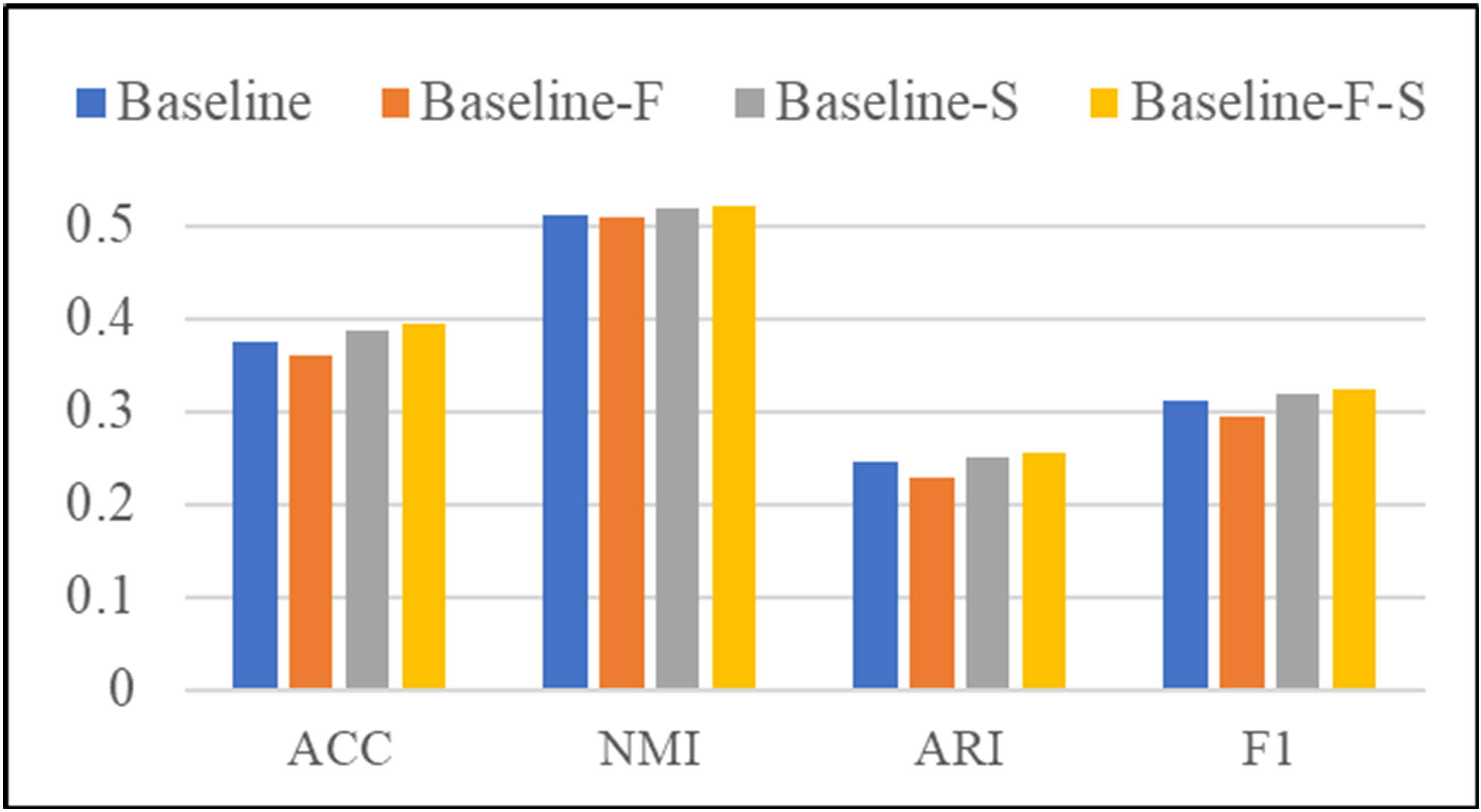}}
\vspace{3pt}
\centerline{CORAFULL}
\end{minipage}
\caption{Ablation comparisons of dual information correlation reduction on six datasets.}
\label{SAMPLE_FEATURE_ABLATION}
\end{figure}

To further investigate the superiority of the proposed DICR mechanism, we experimentally compare our method (i.e., Baseline-F-S in Fig. \ref{SAMPLE_FEATURE_ABLATION}) with three counterparts. Likewise, we denote the DFCN as the Baseline. Baseline-F and Baseline-S are denoted that the Baseline merely adopts feature-level and sample-level correlation reduction strategy, respectively. From the results in Fig. \ref{SAMPLE_FEATURE_ABLATION}, we can see that 1) Baseline-F outperforms Baseline in terms of four matrices on four of six datasets, but obtains unsatisfied performance on DBLP and CORAFULL. This is because the learned embedding is not robust without considering sample-level correlation redundancy; 2) the performance of Baseline-S is consistently better than that of Baseline over all datasets. For instance, Baseline-S obtains 3.60\% accuracy improvement on DBLP. It shows that the decorrelation operation of samples is effective in filtering redundant information of two views while preserving more discriminative features for improving the clustering performance; 3) Baseline-F-S could leverage two types of correlation reduction to make the learned latent embedding more discriminative for better clustering. In summary, the above observations well demonstrate the effectiveness of dual level correlation reduction strategy.



\begin{figure}[!t]
\centering
\tiny
\begin{minipage}{0.48\linewidth}
\centerline{\includegraphics[width=0.9\textwidth]{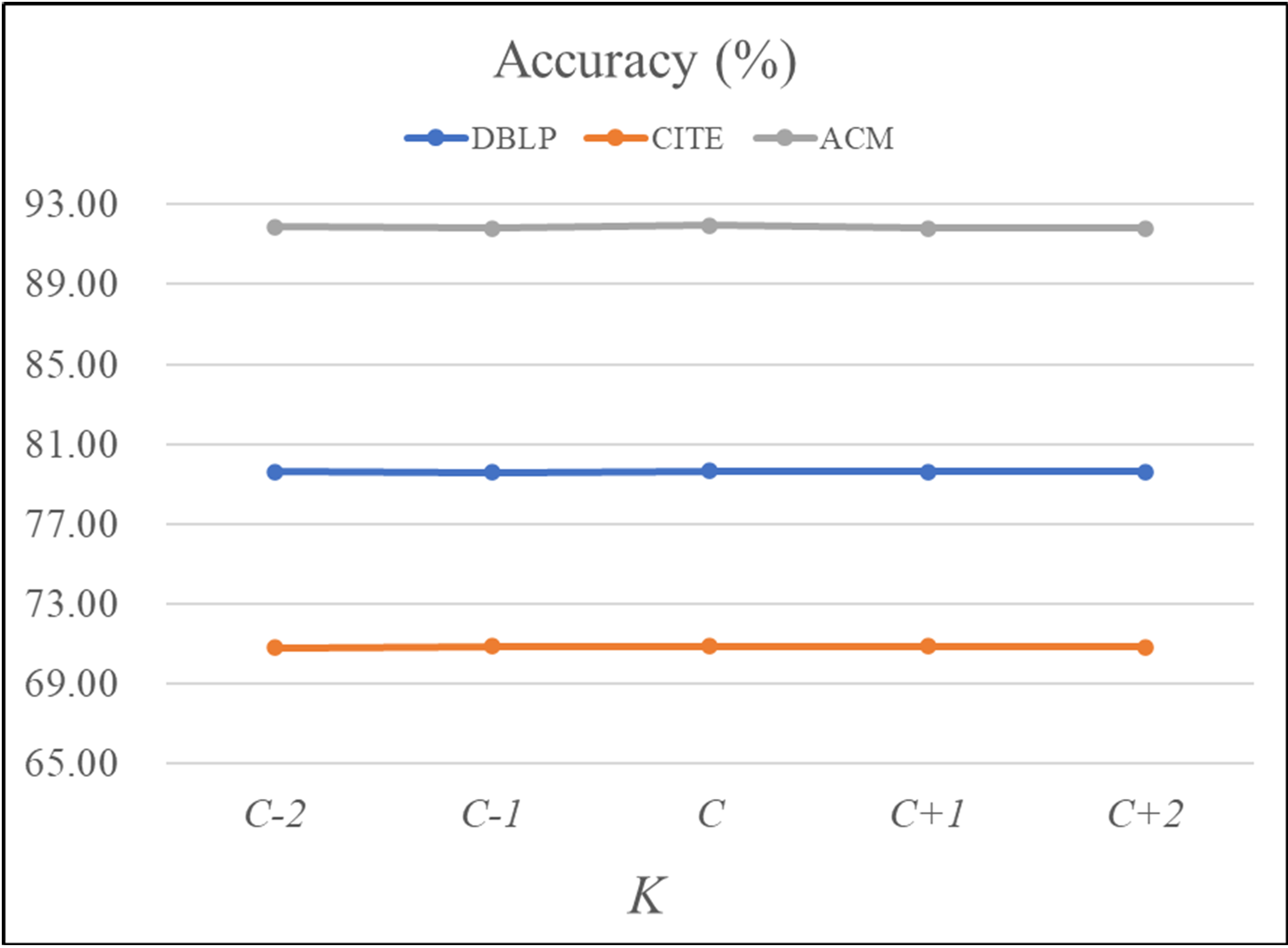}}
\vspace{3pt}

\end{minipage}
\begin{minipage}{0.48\linewidth}
\centerline{\includegraphics[width=0.9\textwidth]{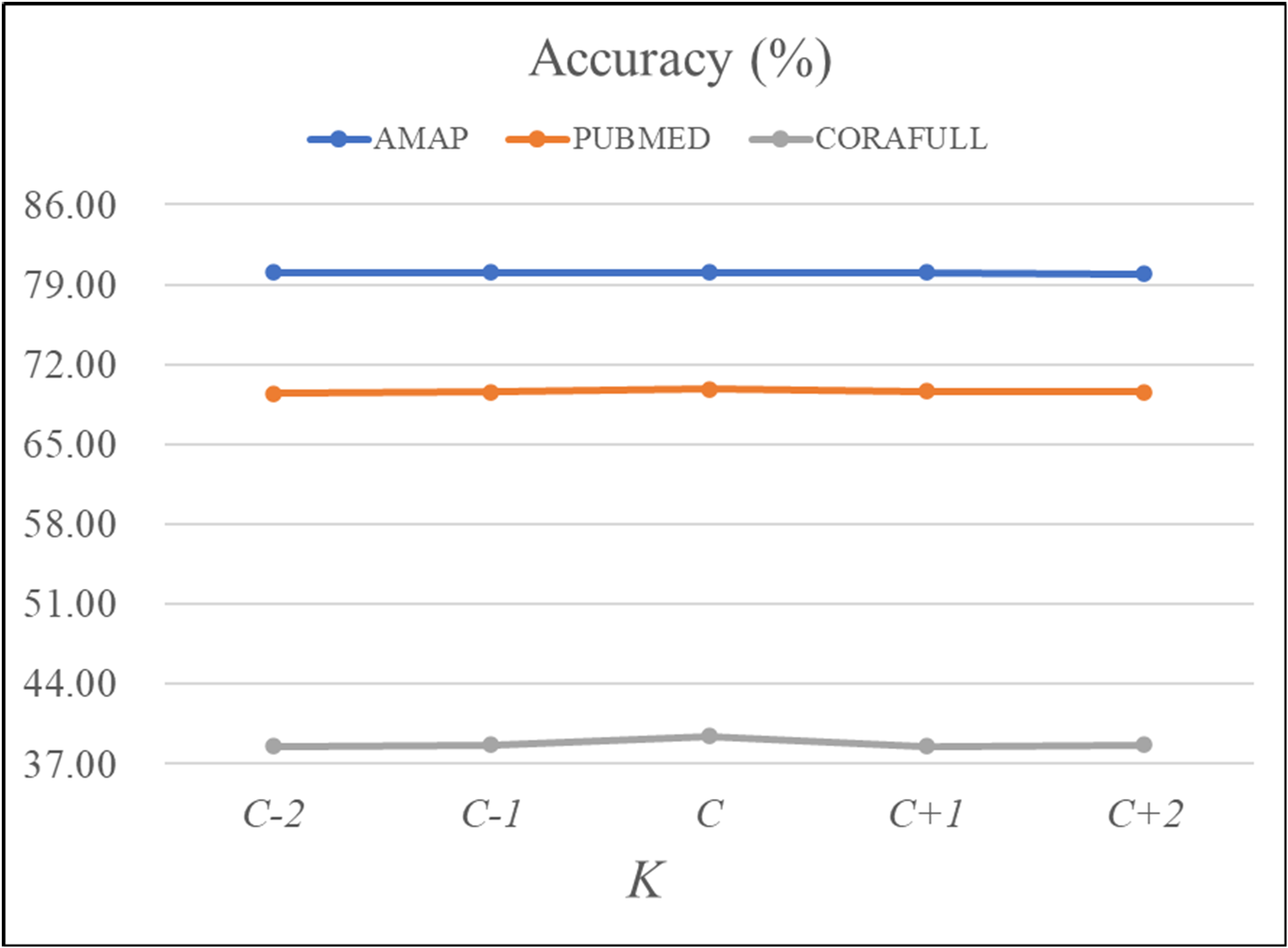}}
\vspace{3pt}

\end{minipage}

\caption{Clustering accuracy vs. hyper-parameter $K$.}
\label{K_ABLATOIN}
\end{figure}

\subsubsection{Hyper-parameter Analysis of $K$}
Furthermore, we investigate the influence of hyper-parameter $K$. From Fig. \ref{K_ABLATOIN}, we observe that 1) the accuracy metric first increases to a high value and generally maintains it up to slight variation with the increasing value $K$; 2) the method tends to perform well when $K$ is equal to the number of clusters $C$; 3) our DCRN is insensitive to the variation of the hyper-parameter $K$.

\subsubsection{$t$-SNE Visualization of Clustering Results}
In order to show the superiority of DRCN intuitively, we visualize the distribution of the learned node embedding $\textbf{Z}$ of DBLP and ACM generated by AE, DEC, GAE, ARGA, DFCN and our DCRN via t-SNE \cite{T_SNE}. As illustrated in Fig. \ref{VIS}, the visual results demonstrate that DCRN have a clearer structure, which can better reveal the intrinsic clustering structure among data.

\section{Conclusion}
In this work, we propose a novel self-supervised deep graph clustering network termed as Dual Correlation Reduction Network (DCRN). In our model, a carefully-designed dual information correlation reduction mechanism is introduced to reduce the information correlation in both sample and feature level. With this mechanism, the redundant information of the latent variables from two views can be filtered out and more discriminative features of both views can be well preserved. It plays an important role in avoiding representation collapse for better clustering. Experimental results on six benchmarks demonstrate the superiority of DCRN. 


\section{Acknowledgments}
This work was supported by the National Key R\&D Program of China (project no. 2020AAA0107100) and the National Natural Science Foundation of China (project no. 61922088, 61906020, 61872371 and 62006237).

\bibliography{ref}

\end{document}